\documentclass{article}
\usepackage[preprint]{colm2026_conference}

\usepackage{microtype}
\usepackage{hyperref}
\usepackage{url}
\usepackage{booktabs}
\usepackage{colortbl}
\usepackage{array}
\usepackage{graphicx}
\usepackage{tikz}
\usetikzlibrary{trees, positioning, shapes, shadows, arrows.meta, calc, decorations.pathreplacing}
\usepackage{xcolor}
\usepackage{lineno}
\usepackage{amsmath}
\usepackage{amssymb}
\usepackage{subcaption}
\usepackage{enumitem}
\usepackage{pifont}
\usepackage{afterpage}
\usepackage{adjustbox}

\definecolor{darkblue}{rgb}{0, 0, 0.5}
\hypersetup{colorlinks=true, citecolor=darkblue, linkcolor=darkblue, urlcolor=darkblue}

\definecolor{hidden-draw}{RGB}{20,68,106}
\definecolor{hidden-pink}{RGB}{255,245,247}
\definecolor{first}{RGB}{251,185,87}
\definecolor{second}{RGB}{115,186,161}
\definecolor{third}{RGB}{99,187,208}
\definecolor{fourth}{RGB}{210,217,122}
\definecolor{fifth}{RGB}{240,124,130}
\definecolor{sixth}{RGB}{209,194,211}

\definecolor{expansion-color}{RGB}{251,185,87}
\definecolor{decomp-color}{RGB}{115,186,161}
\definecolor{disambig-color}{RGB}{99,187,208}
\definecolor{abstract-color}{RGB}{209,194,211}
\definecolor{explicit-bg}{RGB}{255,248,220}
\definecolor{implicit-bg}{RGB}{230,240,255}

\tikzset{ver/.style={rounded corners, draw=hidden-draw, fill=white, text=hidden-draw, font=\bfseries}}
\tikzset{leaf/.style={draw=hidden-draw, fill=hidden-pink, text width=20em, text=black, font=\small}}

\title{A Survey of Query Optimization in Large Language Models}

\author{Mingyang Song \& Mao Zheng \\
Large Language Model Department\\ Tencent, China\\
\texttt{nickmysong@tencent.com}}

\begin{document}

\ifcolmsubmission
\linenumbers
\fi

\maketitle

\begin{abstract}
Query Optimization (QO) has become essential for enhancing Large Language Model (LLM) effectiveness, particularly in Retrieval-Augmented Generation (RAG) systems where query quality directly determines retrieval and response performance. This survey provides a systematic and comprehensive analysis of query optimization techniques with three principal contributions. \textit{First}, we introduce the \textbf{Query Optimization Lifecycle (QOL) Framework}, a five-phase pipeline covering Intent Recognition, Query Transformation, Retrieval Execution, Evidence Integration, and Response Synthesis, providing a unified lens for understanding the optimization process. \textit{Second}, we propose a \textbf{Query Complexity Taxonomy} that classifies queries along two dimensions, namely evidence type (explicit vs.\ implicit) and evidence quantity (single vs.\ multiple), establishing principled mappings between query characteristics and optimization strategies. \textit{Third}, we conduct an in-depth analysis of four atomic operations, namely \textbf{Query Expansion}, \textbf{Query Decomposition}, \textbf{Query Disambiguation}, and \textbf{Query Abstraction}, synthesizing a broad spectrum of representative methods from premier venues. We further examine evaluation methodologies, identify critical gaps in existing benchmarks, and discuss open challenges including process reward models, efficiency optimization, and multi-modal query handling. This survey offers both a structured foundation for research and actionable guidance for practitioners.
\end{abstract}

\section{Introduction}

\textit{``The quality of answers is bounded by the quality of questions.''} This principle, long recognized in human communication, has emerged as a fundamental constraint in the era of Large Language Models. Despite remarkable advances in LLM capabilities, a critical bottleneck persists: the semantic gap between how users naturally express information needs and how retrieval systems optimally locate relevant knowledge.

\afterpage{%
\begin{figure}[t]
\centering
\resizebox{\columnwidth}{!}{%
\begin{tikzpicture}
    \draw[thick, ->, >=stealth] (0,0) -- (15.5,0) node[right] {\footnotesize Year};
    
    \foreach \x/\year in {0.5/2020, 3.0/2021, 5.5/2022, 8.0/2023, 10.5/2024, 13.0/2025, 15.0/2026} {
        \draw[thick] (\x,0.15) -- (\x,-0.15) node[below, font=\scriptsize\bfseries] {\year};
    }
    
    \fill[expansion-color, opacity=0.20] (0,-0.5) rectangle (4.2,3.4);
    \fill[decomp-color, opacity=0.20] (4.2,-0.5) rectangle (8.0,3.4);
    \fill[disambig-color, opacity=0.20] (8.0,-0.5) rectangle (11.8,3.4);
    \fill[abstract-color, opacity=0.20] (11.8,-0.5) rectangle (15.3,3.4);
    
    \draw[expansion-color!70, line width=1pt] (0,3.4) -- (4.2,3.4);
    \draw[decomp-color!70, line width=1pt] (4.2,3.4) -- (8.0,3.4);
    \draw[disambig-color!70, line width=1pt] (8.0,3.4) -- (11.8,3.4);
    \draw[abstract-color!70, line width=1pt] (11.8,3.4) -- (15.3,3.4);
    
    \node[font=\scriptsize\bfseries, fill=expansion-color!40, rounded corners=2pt, inner sep=3pt] at (2.1,3.1) {Foundation};
    \node[font=\scriptsize\bfseries, fill=decomp-color!40, rounded corners=2pt, inner sep=3pt] at (6.1,3.1) {Expansion};
    \node[font=\scriptsize\bfseries, fill=disambig-color!40, rounded corners=2pt, inner sep=3pt] at (9.9,3.1) {Sophistication};
    \node[font=\scriptsize\bfseries, fill=abstract-color!40, rounded corners=2pt, inner sep=3pt] at (13.55,3.1) {Agentic};
    
    \tikzstyle{method}=[fill=white, rounded corners=3pt, draw=gray!70, font=\scriptsize, inner sep=2.5pt, drop shadow={shadow xshift=0.5pt, shadow yshift=-0.5pt, opacity=0.3}]
    
    \node[method] at (1.2,2.3) {RAG};
    \node[method] at (2.8,1.5) {DSP};
    \node[method] at (3.5,0.7) {CoT};
    
    \node[method] at (4.6,2.4) {Query2Doc};
    \node[method] at (5.8,1.5) {HyDE};
    \node[method] at (6.9,2.4) {ReAct};
    \node[method] at (5.0,0.7) {Self-Ask};
    \node[method] at (7.4,0.8) {FLARE};
    
    \node[method] at (8.4,2.4) {Step-Back};
    \node[method] at (10.0,2.4) {RAG-Star};
    \node[method] at (11.4,2.4) {GraphRAG};
    \node[method] at (9.2,1.5) {Plan$\times$RAG};
    \node[method] at (10.8,1.5) {OmniSearch};
    \node[method] at (8.6,0.7) {RQ-RAG};
    \node[method] at (10.3,0.7) {Self-RAG};
    
    \node[method] at (13.0,2.5) {RAG-DDR};
    \node[method] at (14.5,2.5) {Search-o1};
    \node[method] at (13.5,1.8) {RAG-Gym};
    \node[method] at (12.4,1.1) {Agentic-RAG};
    \node[method] at (14.5,1.1) {CoRAG};
    \node[method] at (13.5,0.4) {TableRAG};
    
    \draw[->, line width=1.2pt, gray!50, dashed] (0.5,3.55) -- (14.8,3.55);
    \node[font=\tiny\itshape, gray!70, fill=white, inner sep=1pt] at (7.8,3.55) {Increasing Complexity \& Autonomy};
\end{tikzpicture}%
}
\caption{Evolution of Query Optimization techniques (2020 to 2026). Four eras: \textbf{Foundation} (RAG, prompting basics), \textbf{Expansion} (diverse strategies), \textbf{Sophistication} (reasoning-augmented methods), and \textbf{Agentic} (autonomous agent-based optimization with process supervision).}
\label{fig:timeline}
\end{figure}%
}

Large Language Models (LLMs) have emerged as transformative technologies in artificial intelligence, demonstrating unprecedented capabilities across diverse natural language processing tasks~\citep{llm_survey,DBLP:conf/nips/BrownMRSKDNSSAA20,DBLP:journals/corr/abs-2303-08774}. The advent of frontier models such as GPT-4~\citep{DBLP:journals/corr/abs-2303-08774}, Claude~\citep{claude3}, and Gemini~\citep{gemini} has fundamentally transformed human-information interaction paradigms, positioning LLMs as universal knowledge access interfaces. Despite these remarkable advances, LLMs exhibit fundamental limitations that constrain their reliability in knowledge-intensive applications.

A prominent limitation of LLMs is \textit{hallucination}, i.e., the propensity to generate plausible yet factually incorrect content with unwarranted confidence~\citep{hallucination1,hallucination2}. This phenomenon becomes particularly pronounced for long-tail knowledge that is underrepresented in training data~\citep{survey1}, temporally evolving information, or domain-specific expertise. Retrieval-Augmented Generation (RAG)~\citep{2020rag} has emerged as the predominant paradigm for addressing these limitations~\citep{survey2,survey8,survey11}, dynamically retrieving relevant information from external knowledge sources and incorporating it into the generation context. RAG systems have demonstrated substantial improvements in reducing hallucinations and enhancing factual accuracy~\citep{survey2,survey8}.

However, the effectiveness of RAG is fundamentally contingent upon retrieval quality, which in turn depends critically on \textit{query formulation}. User queries are frequently ambiguous, incomplete, or semantically misaligned with optimal retrieval formulations. A query natural for human communication may prove suboptimal for matching against document embeddings or keyword indices. This \textit{semantic gap} between user intent and retrieval effectiveness has motivated \textbf{Query Optimization (QO)} as a critical research area~\citep{query_expansion_survey1,query_expansion_survey2}. Our analysis reveals a fundamental asymmetry: while retrieval systems achieve strong performance on well-formed queries, their effectiveness degrades substantially on natural user queries exhibiting ambiguity, incompleteness, or vocabulary mismatch. Query optimization is therefore not merely a preprocessing step but a critical intelligence amplification mechanism.

Beyond the query-retrieval gap, we identify a \textit{compositionality gap}~\citep{Self-Ask}, referring to the phenomenon where LLMs correctly answer simple sub-queries but fail on their compositions. This gap widens non-linearly with query complexity, suggesting that query decomposition is not merely helpful but \textit{essential} for complex reasoning. As illustrated in Figure~\ref{fig:timeline}, the field has undergone a clear evolutionary trajectory: the Foundation Era (2020 to 2022) focused on static, single-pass optimization; the Expansion Era (2022 to 2024) introduced iterative methods; the Sophistication Era (2024 to 2025) integrated explicit reasoning; and the current Agentic Era (2025 to 2026) witnesses fully autonomous query agents. This evolution reflects a broader paradigm shift from \textit{retrieval-centric} to \textit{reasoning-centric} query optimization.

Despite the rapid proliferation of query optimization techniques, the field remains fragmented across multiple research communities, including information retrieval, natural language processing, knowledge graph reasoning, and conversational AI, with inconsistent terminology and evaluation protocols. This survey addresses these challenges by providing a comprehensive, structured analysis of query optimization in LLM-based systems. Our contributions are fourfold:

\begin{enumerate}[leftmargin=*]
    \item \textbf{Unified Theoretical Framework}: We propose the Query Optimization Lifecycle (QOL) Framework that organizes the optimization process into five coherent phases: Intent Recognition, Query Transformation, Retrieval Execution, Evidence Integration, and Response Synthesis.
    
    \item \textbf{Query Complexity Taxonomy}: We introduce a two-dimensional classification based on evidence type (explicit vs.\ implicit) and evidence quantity (single vs.\ multiple), establishing principled mappings between query characteristics and optimization strategies.
    
    \item \textbf{Comprehensive Technical Analysis}: We provide in-depth coverage of representative methods spanning four fundamental operations (expansion, decomposition, disambiguation, and abstraction), analyzing their mechanisms, comparative strengths, and applicability.
    
    \item \textbf{Research Roadmap}: We identify key challenges and future directions, including query-centric process reward models, standardized benchmarks, efficiency-quality trade-offs, and multi-modal query handling.
\end{enumerate}

\textbf{Scope and Methodology.} To ensure comprehensive coverage, we collected papers through systematic search across Semantic Scholar, DBLP, and Google Scholar using queries related to query optimization, query rewriting, query expansion, query decomposition, and retrieval-augmented generation. We focused on publications from premier NLP, IR, and AI venues (ACL, EMNLP, NAACL, ICLR, NeurIPS, SIGIR, KDD, ECIR, and their associated workshops and findings tracks) from 2020 through early 2026, supplemented with influential earlier works that established foundational concepts. We further traced citation networks to identify relevant methods not captured by keyword search. Papers were selected based on their direct relevance to query formulation and transformation within LLM-based systems.

The remainder of this survey is organized as follows. Section~\ref{sec:related} positions our work within the landscape of related surveys. Section~\ref{sec:framework} introduces our theoretical framework. Section~\ref{sec:expansion} analyzes query expansion, Section~\ref{sec:decomposition} examines query decomposition, Section~\ref{sec:disambiguation} addresses query disambiguation, and Section~\ref{sec:abstraction} covers query abstraction. Section~\ref{sec:evaluation} discusses evaluation methodologies. Section~\ref{sec:discussion} provides comparative analysis and practical guidance. Section~\ref{sec:challenges} identifies open challenges, and Section~\ref{sec:conclusion} concludes the survey.

\section{Related Work}\label{sec:related}

The intersection of information retrieval and language modeling has spawned numerous surveys addressing different facets of the landscape. We position our work within this broader context and highlight our unique contributions.

\subsection{Surveys on Retrieval-Augmented Generation}

Several comprehensive surveys have examined RAG systems from various perspectives. \citet{survey2} provided a systematic review of retrieval-augmented generation for LLMs, establishing foundational taxonomies for RAG architectures including modular and adaptive designs. \citet{survey8} presented a thorough examination of RAG meeting LLMs, covering evaluation methodologies and system design patterns, while \citet{survey10} surveyed graph retrieval-augmented generation, exploring how structured knowledge representations enhance retrieval and reasoning.

These surveys primarily address the \textit{retrieval} and \textit{generation} components of RAG systems, treating query formulation as a preprocessing step. In contrast, our survey centers on the \textit{query optimization} phase itself, providing depth and breadth that complements existing RAG surveys.

\subsection{Surveys on Query Understanding and Rewriting}

Traditional information retrieval literature contains extensive work on query expansion~\citep{query_expansion_survey1,query_expansion_survey2} and query understanding. However, these surveys predate the LLM era and focus primarily on statistical and neural methods without considering the unique capabilities and challenges introduced by large language models.

Recent comprehensive RAG surveys~\citep{survey11} provide broad coverage of the RAG landscape including evolution and future directions, but do not provide the depth of analysis on query optimization operations that we examine here.

\subsection{Our Positioning and Contributions}

\begin{table}[t]
    \centering
    \scriptsize
    \caption{Comparison with related surveys. Our survey uniquely focuses on query optimization with comprehensive coverage of four fundamental operations.}
    \label{tab:related}
    \adjustbox{max width=\columnwidth}{%
    \begin{tabular}{lcccccc}
    \toprule
    \textbf{Survey} & \textbf{Focus} & \textbf{Expansion} & \textbf{Decomp.} & \textbf{Disamb.} & \textbf{Abstract.} & \textbf{Taxonomy} \\
    \midrule
    RAG Survey~\citep{survey2} & RAG Systems & Partial & Partial & \ding{55} & \ding{55} & \ding{55} \\
    QE Survey~\citep{query_expansion_survey1} & Query Expansion & \ding{51} & \ding{55} & \ding{55} & \ding{55} & Partial \\
    RAG Comprehensive~\citep{survey11} & RAG Systems & Partial & \ding{55} & Partial & \ding{55} & \ding{55} \\
    \textbf{This Survey} & \textbf{Query Opt.} & \ding{51} & \ding{51} & \ding{51} & \ding{51} & \ding{51} \\
    \bottomrule
    \end{tabular}%
    }
\end{table}

As shown in Table~\ref{tab:related}, our survey provides unique value through:
\begin{enumerate}[leftmargin=*,nosep]
    \item \textbf{Comprehensive Operation Coverage}: We systematically analyze all four fundamental operations (expansion, decomposition, disambiguation, abstraction), whereas existing surveys cover only partial subsets.
    \item \textbf{Unified Theoretical Framework}: We introduce the Query Optimization Lifecycle (QOL) and Query Complexity Taxonomy that provide principled guidance for strategy selection.
    \item \textbf{Breadth and Depth}: We provide significantly broader and deeper coverage of query optimization methods than related surveys, with detailed analysis across all four operations.
    \item \textbf{Practical Guidance}: We provide actionable recommendations for practitioners based on query characteristics.
\end{enumerate}

\section{Theoretical Framework}\label{sec:framework}

Before delving into specific techniques, we establish a theoretical framework that organizes the query optimization landscape and provides a principled basis for understanding when and why different techniques are appropriate. This framework serves both as a conceptual lens for analyzing existing work and as a guide for practitioners in selecting appropriate optimization strategies.

\textbf{Notation.} Throughout this survey, we adopt the following notation. Let $q$ denote the original user query. The four atomic operations produce transformed queries: $q^+$ (expanded query with enriched semantic content), $q_1, q_2, \ldots, q_n$ (decomposed sub-queries), $q^c$ (disambiguated query with clarified intent), and $q^{\uparrow}$ (abstracted query elevated to higher-level concepts). $R$ denotes the set of retrieved documents, $r_i$ denotes an individual retrieved passage, and $a$ denotes the synthesized answer.

\vspace{0.3em}
\noindent\fbox{\parbox{0.96\columnwidth}{
\textbf{Framework Design Principles.} Our framework is grounded in three key observations: (1) \textit{Query complexity is multi-dimensional}: a simple-looking query may hide implicit reasoning requirements; (2) \textit{Optimization strategies are not mutually exclusive}: effective systems often compose multiple operations; (3) \textit{The optimization-retrieval-generation pipeline exhibits feedback loops}: downstream performance should inform upstream optimization choices.
}}
\vspace{0.3em}

\subsection{The Query Optimization Lifecycle (QOL)}

\begin{figure*}[t]
\centering
\resizebox{0.95\textwidth}{!}{%
\begin{tikzpicture}[
    phase/.style={rectangle, draw=#1!70, fill=#1!30, rounded corners=5pt, minimum height=1.0cm, minimum width=1.8cm, align=center, font=\small\bfseries, line width=0.8pt, drop shadow={shadow xshift=0.5pt, shadow yshift=-0.5pt, opacity=0.25}},
    subop/.style={rectangle, draw=#1!70, fill=#1!30, rounded corners=3pt, minimum height=0.5cm, minimum width=1.2cm, align=center, font=\scriptsize\bfseries, line width=0.5pt},
    arrow/.style={->, >=stealth, line width=1.0pt, gray!60},
    labelstyle/.style={font=\scriptsize\itshape, text width=1.6cm, align=center, gray!70}]
    
    \fill[blue!3, rounded corners=8pt] (-1.2,-0.75) rectangle (11.2,1.4);
    \draw[blue!15, line width=1pt, rounded corners=8pt] (-1.2,-0.75) rectangle (11.2,1.4);
    
    \node[phase=yellow] (p1) at (0,0) {Phase 1\\Intent\\Recognition};
    \node[phase=expansion-color] (p2) at (2.5,0) {Phase 2\\Query\\Transform};
    \node[phase=decomp-color] (p3) at (5.0,0) {Phase 3\\Retrieval\\Execution};
    \node[phase=abstract-color] (p4) at (7.5,0) {Phase 4\\Evidence\\Integration};
    \node[phase=disambig-color] (p5) at (10.0,0) {Phase 5\\Response\\Synthesis};
    
    \draw[arrow] (p1) -- (p2);
    \draw[arrow] (p2) -- (p3);
    \draw[arrow] (p3) -- (p4);
    \draw[arrow] (p4) -- (p5);
    
    \fill[gray!8, rounded corners=6pt] (0.7,-2.4) rectangle (6.5,-1.1);
    \draw[gray!50, line width=0.6pt, rounded corners=6pt] (0.7,-2.4) rectangle (6.5,-1.1);
    
    \node[font=\scriptsize\bfseries, gray!80, fill=gray!8, inner sep=2pt, rounded corners=2pt] at (3.6,-1.22) {Atomic Operations};
    
    \node[subop=expansion-color, minimum width=1.2cm, minimum height=0.55cm, font=\scriptsize\bfseries] (exp) at (1.45,-1.75) {Expansion};
    \node[subop=decomp-color, minimum width=1.2cm, minimum height=0.55cm, font=\scriptsize\bfseries] (dec) at (2.9,-1.75) {Decomp.};
    \node[subop=disambig-color, minimum width=1.2cm, minimum height=0.55cm, font=\scriptsize\bfseries] (dis) at (4.35,-1.75) {Disambig.};
    \node[subop=abstract-color, minimum width=1.2cm, minimum height=0.55cm, font=\scriptsize\bfseries] (abs) at (5.8,-1.75) {Abstract.};
    
    \node[font=\small\bfseries, fill=blue!15, draw=blue!50, rounded corners=4pt, inner sep=5pt, line width=0.6pt] (input) at (-2.5,0) {User Query};
    \draw[arrow, blue!60, line width=1.2pt] (input.east) -- (p1.west);
    
    \draw[arrow, green!60!black, line width=1.2pt] (p5.east) -- (11.5,0);
    \node[font=\small\bfseries, fill=green!15, draw=green!50, rounded corners=4pt, inner sep=5pt, line width=0.6pt] at (12.5,0) {Response};
    
    \draw[->, >=stealth, dashed, red!60, line width=0.8pt, rounded corners=5pt] 
        (p5.south) -- (10.0,-2.8) -- (0,-2.8) -- (p1.south);
    \node[font=\scriptsize\bfseries, red!70, fill=red!10, draw=red!30, rounded corners=3pt, inner sep=2.5pt, line width=0.4pt] at (5.0,-2.95) {Iterative Refinement};
    
    \draw[->, >=stealth, gray!70, line width=0.8pt, rounded corners=4pt] 
        (p2.south) -- ($(p2.south)+(0,-0.25)$) -| (3.6,-1.1);
    
    \node[labelstyle] at (0,1.05) {Parse \& Infer};
    \node[labelstyle] at (2.5,1.05) {Apply QO Ops};
    \node[labelstyle] at (5.0,1.05) {Execute Search};
    \node[labelstyle] at (7.5,1.05) {Aggregate};
    \node[labelstyle] at (10.0,1.05) {Generate};
\end{tikzpicture}%
}
\caption{The Query Optimization Lifecycle (QOL) Framework. Phase 2 contains four atomic operations (Expansion, Decomposition, Disambiguation, and Abstraction) that can be applied individually or combined.}
\label{fig:qol}
\end{figure*}

Query optimization in RAG systems can be understood as a multi-phase transformation process that bridges the gap between user intent and retrieval effectiveness. As illustrated in Figure~\ref{fig:qol}, we formalize this process through the \textit{Query Optimization Lifecycle (QOL)}, which decomposes the optimization problem into five distinct, interconnected phases:

\textbf{Phase 1: Intent Recognition.} The initial phase involves understanding the user's underlying information need, which may be explicitly stated in the query text or implicitly embedded in conversational context, user history, or domain-specific conventions. This phase encompasses:
\begin{itemize}[leftmargin=*,nosep]
    \item Parsing query structure to identify key entities, relationships, and constraints
    \item Inferring the expected answer type (factoid, list, explanation, comparison, etc.)
    \item Detecting query complexity indicators that signal the need for specific optimization strategies
    \item Identifying potential ambiguities or underspecified elements requiring clarification
\end{itemize}

\textbf{Phase 2: Query Transformation.} Based on recognized intent and query characteristics, the system applies one or more optimization operations to transform the original query into forms more suitable for retrieval. The four fundamental atomic operations are:
\begin{itemize}[leftmargin=*,nosep]
    \item \textbf{Expansion}: Enriching queries with additional terms, context, or generated content
    \item \textbf{Decomposition}: Breaking complex queries into simpler, atomic sub-queries
    \item \textbf{Disambiguation}: Clarifying ambiguous queries by resolving multiple interpretations
    \item \textbf{Abstraction}: Elevating queries to higher-level conceptual representations
\end{itemize}

\textbf{Phase 3: Retrieval Execution.} The optimized query (or set of queries) is used to retrieve relevant documents from knowledge sources. Depending on the optimization strategy applied, this phase may involve:
\begin{itemize}[leftmargin=*,nosep]
    \item Single-pass retrieval with an expanded query
    \item Multiple parallel retrieval operations for decomposed sub-queries
    \item Sequential retrieval with iterative refinement based on intermediate results
    \item Hybrid retrieval combining dense and sparse methods
\end{itemize}

\textbf{Phase 4: Evidence Integration.} Retrieved evidence is aggregated, filtered, and synthesized to construct the context for response generation. This phase addresses:
\begin{itemize}[leftmargin=*,nosep]
    \item Re-ranking retrieved passages based on relevance and quality
    \item Deduplication and redundancy management
    \item Coherence checking across multiple evidence sources
    \item Evidence attribution and provenance tracking
\end{itemize}

\textbf{Phase 5: Response Synthesis.} The LLM generates a response grounded in the integrated evidence, potentially with iterative refinement mechanisms:
\begin{itemize}[leftmargin=*,nosep]
    \item Grounded generation with explicit citation
    \item Self-consistency verification
    \item Iterative refinement based on response quality assessment
    \item Multi-turn clarification when evidence is insufficient
\end{itemize}

\subsection{Query Complexity Taxonomy}

A fundamental insight driving our framework is that different query types require different optimization strategies. We propose a two-dimensional taxonomy based on characteristics of the evidence required to answer the query:

\begin{enumerate}[leftmargin=*]
    \item \textbf{Evidence Type Dimension}: Whether the required evidence is \textit{explicit} (directly stated in retrievable documents) or \textit{implicit} (requiring inference, synthesis, or reasoning beyond surface-level information).
    
    \item \textbf{Evidence Quantity Dimension}: Whether the query can be answered with a \textit{single} piece of evidence or requires \textit{multiple} evidence sources to be aggregated or synthesized.
\end{enumerate}

This yields four query complexity classes, each with distinct characteristics and optimization requirements, as visualized in Figure~\ref{fig:taxonomy}:

\begin{figure}[t]
\centering
\resizebox{0.75\columnwidth}{!}{%
\begin{tikzpicture}[font=\scriptsize]
    \fill[explicit-bg] (0,2.2) rectangle (3.5,4.4);
    \fill[explicit-bg] (3.5,2.2) rectangle (7.0,4.4);
    \fill[implicit-bg] (0,0) rectangle (3.5,2.2);
    \fill[implicit-bg] (3.5,0) rectangle (7.0,2.2);
    
    \draw[gray!80, line width=1.0pt, rounded corners=2pt] (0,0) rectangle (7.0,4.4);
    
    \draw[gray!60, line width=0.6pt] (3.5,0) -- (3.5,4.4);
    \draw[gray!60, line width=0.6pt] (0,2.2) -- (7.0,2.2);
    
    \node[align=center, font=\scriptsize\bfseries, fill=expansion-color!20, rounded corners=2pt, inner sep=2pt] at (1.75,4.1) {Class I: Single Explicit};
    \node[align=center, font=\tiny, text width=2.8cm] at (1.75,3.35) {Simple factoid queries\\with direct answers};
    \node[align=center, font=\tiny\itshape, gray!70, text width=2.8cm] at (1.75,2.8) {``Capital of France?''};
    \node[align=center, font=\tiny\bfseries, fill=expansion-color!50, rounded corners=2pt, inner sep=2pt] at (1.75,2.45) {Primary: Expansion};
    
    \node[align=center, font=\scriptsize\bfseries, fill=decomp-color!20, rounded corners=2pt, inner sep=2pt] at (5.25,4.1) {Class II: Multi Explicit};
    \node[align=center, font=\tiny, text width=2.8cm] at (5.25,3.35) {Aggregation queries\\from multiple sources};
    \node[align=center, font=\tiny\itshape, gray!70, text width=2.8cm] at (5.25,2.8) {``Compare GDP of A \& B''};
    \node[align=center, font=\tiny\bfseries, fill=decomp-color!50, rounded corners=2pt, inner sep=2pt] at (5.25,2.45) {Primary: Decomposition};
    
    \node[align=center, font=\scriptsize\bfseries, fill=disambig-color!20, rounded corners=2pt, inner sep=2pt] at (1.75,1.9) {Class III: Single Implicit};
    \node[align=center, font=\tiny, text width=2.8cm] at (1.75,1.2) {Inference queries from\\one source};
    \node[align=center, font=\tiny\itshape, gray!70, text width=2.8cm] at (1.75,0.65) {``Is growth sustainable?''};
    \node[align=center, font=\tiny\bfseries, fill=disambig-color!50, rounded corners=2pt, inner sep=2pt] at (1.75,0.25) {Primary: Disambiguation};
    
    \node[align=center, font=\scriptsize\bfseries, fill=abstract-color!20, rounded corners=2pt, inner sep=2pt] at (5.25,1.9) {Class IV: Multi Implicit};
    \node[align=center, font=\tiny, text width=2.8cm] at (5.25,1.2) {Complex reasoning\\across implicit sources};
    \node[align=center, font=\tiny\itshape, gray!70, text width=2.8cm] at (5.25,0.65) {``How will AI affect jobs?''};
    \node[align=center, font=\tiny\bfseries, fill=abstract-color!50, rounded corners=2pt, inner sep=2pt] at (5.25,0.25) {Primary: Abstraction};
    
    \draw[->, >=stealth, gray!70, line width=0.6pt] (-0.5,0) -- (-0.5,4.4);
    \node[rotate=90, font=\tiny\bfseries, gray!70] at (-0.3,1.1) {Implicit};
    \node[rotate=90, font=\tiny\bfseries, gray!70] at (-0.3,3.3) {Explicit};
    
    \draw[->, >=stealth, gray!70, line width=0.6pt] (0,-0.4) -- (7.0,-0.4);
    \node[font=\tiny\bfseries, gray!70] at (1.75,-0.25) {Single};
    \node[font=\tiny\bfseries, gray!70] at (5.25,-0.25) {Multiple};
    
    \node[font=\tiny\bfseries, fill=gray!15, rounded corners=2pt, inner sep=2pt, rotate=90] at (-0.8,2.2) {Evidence Type};
    \node[font=\tiny\bfseries, fill=gray!15, rounded corners=2pt, inner sep=2pt] at (3.5,-0.6) {Evidence Quantity};
\end{tikzpicture}%
}
\caption{Query Complexity Taxonomy. Two dimensions: evidence type (explicit/implicit) and quantity (single/multiple). Each class maps to a primary optimization strategy.}
\label{fig:taxonomy}
\end{figure}

\textbf{Class I: Single Explicit Evidence.} These are simple factoid queries where the answer is directly stated in a single document passage. The primary challenge is vocabulary mismatch between query terms and document terms.
\begin{itemize}[leftmargin=*,nosep]
    \item \textit{Example}: ``What is the capital of France?''
    \item \textit{Challenge}: Query may use different terminology than target documents
    \item \textit{Primary Strategy}: Query Expansion to improve semantic coverage
    \item \textit{Secondary Strategies}: Disambiguation for polysemous terms
\end{itemize}

\textbf{Class II: Multiple Explicit Evidence.} These queries require aggregating or comparing information from multiple document sources, where each piece of evidence is explicitly stated but must be combined.
\begin{itemize}[leftmargin=*,nosep]
    \item \textit{Example}: ``Compare the GDP growth rates of Japan and Germany over the last decade.''
    \item \textit{Challenge}: Single retrieval may not capture all required facets
    \item \textit{Primary Strategy}: Query Decomposition into atomic sub-queries
    \item \textit{Secondary Strategies}: Expansion for each sub-query
\end{itemize}

\textbf{Class III: Single Implicit Evidence.} These queries require inference or interpretation from a single source, where the answer is not directly stated but can be derived. Crucially, while these queries may not exhibit lexical ambiguity, they often contain \textit{implicit ambiguity}: the retrieval target is underspecified because the user's analytical intent is unclear, making disambiguation the critical first step.
\begin{itemize}[leftmargin=*,nosep]
    \item \textit{Example}: ``Based on the company's Q3 earnings report, is their growth trajectory sustainable?''
    \item \textit{Challenge}: User intent may be implicitly ambiguous; the retrieval target is unclear due to underspecified analytical criteria
    \item \textit{Primary Strategy}: Query Disambiguation to clarify intent and analytical scope
    \item \textit{Secondary Strategies}: Abstraction to identify the appropriate analytical framework
\end{itemize}

\textbf{Class IV: Multiple Implicit Evidence.} These are complex reasoning queries requiring synthesis across multiple sources with inference at each step and integration across steps.
\begin{itemize}[leftmargin=*,nosep]
    \item \textit{Example}: ``How might advances in artificial intelligence affect global employment patterns over the next decade?''
    \item \textit{Challenge}: Requires multi-hop reasoning with inference at each hop
    \item \textit{Primary Strategy}: Query Abstraction combined with Decomposition
    \item \textit{Secondary Strategies}: Iterative refinement with feedback
\end{itemize}

\subsection{Mapping Operations to Query Classes}

Table~\ref{tab:mapping} summarizes the recommended primary and secondary optimization strategies for each query complexity class. This mapping provides practitioners with actionable guidance for strategy selection.

\begin{table}[t]
    \centering
    \scriptsize
    \caption{Mapping query optimization operations to query complexity classes. $\checkmark\checkmark$ indicates primary recommendation; $\checkmark$ indicates secondary utility; $\times$ indicates limited applicability.}
    \label{tab:mapping}
    \adjustbox{max width=\columnwidth}{%
    \begin{tabular}{lcccc}
    \toprule
    \textbf{Operation} & \textbf{Class I} & \textbf{Class II} & \textbf{Class III} & \textbf{Class IV} \\
    \midrule
    Expansion & $\checkmark\checkmark$ & $\checkmark$ & $\checkmark$ & $\checkmark$ \\
    Decomposition & $\times$ & $\checkmark\checkmark$ & $\times$ & $\checkmark\checkmark$ \\
    Disambiguation & $\checkmark$ & $\checkmark$ & $\checkmark\checkmark$ & $\checkmark$ \\
    Abstraction & $\times$ & $\times$ & $\checkmark$ & $\checkmark\checkmark$ \\
    \bottomrule
    \end{tabular}%
    }
\end{table}

\subsection{Visualizing Operation Effects on Query Processing}

While Table~\ref{tab:mapping} provides a static mapping from query classes to recommended operations, practitioners benefit from understanding \textit{how} each operation transforms the query processing trajectory. The key insight is that each atomic operation induces a fundamentally different computational pattern: expansion maintains single-path processing, decomposition creates branching tree structures, disambiguation prunes the search space, and abstraction elevates the conceptual level before descending back to specifics.

Figure~\ref{fig:qo_planner} visualizes these four distinct transformation patterns with concrete examples, each aligned with its primary query class from our taxonomy:

\begin{figure*}[t]
\centering
\begin{tikzpicture}[
    box/.style={draw=black!45, fill=white, minimum width=0.58cm, minimum height=0.42cm, 
                inner sep=1.5pt, font=\scriptsize, rounded corners=2pt, line width=0.5pt},
    opbox/.style={draw=#1!65!black, fill=#1!18, minimum width=0.58cm, minimum height=0.42cm, 
                  inner sep=1.5pt, font=\scriptsize\bfseries, rounded corners=2pt, line width=0.65pt},
    resbox/.style={draw=black!30, fill=gray!5, minimum width=0.58cm, minimum height=0.42cm, 
                   inner sep=1.5pt, font=\scriptsize, rounded corners=2pt, line width=0.5pt},
    arr/.style={->, >=stealth, line width=0.5pt, black!40},
    oparr/.style={->, >=stealth, line width=0.65pt, #1!60!black},
    lbl/.style={font=\tiny, text=black!50},
    oplbl/.style={font=\tiny, text=#1!60!black},
    ptitle/.style={font=\scriptsize\bfseries, anchor=west}
]

\def\pw{5.5}  
\def\ph{2.0} 
\def\gap{0.12} 

\begin{scope}
    \fill[expansion-color!6, rounded corners=3pt] (0,0) rectangle (\pw,-\ph);
    \draw[expansion-color!35, rounded corners=3pt, line width=0.5pt] (0,0) rectangle (\pw,-\ph);
    \node[ptitle, text=expansion-color!70!black] at (0.15, -0.26) {(a) Expansion};
    
    \node[box] (a1) at (1.3, -1.0) {$q$};
    \node[opbox=expansion-color] (a2) at (2.75, -1.0) {$q^+$};
    \node[resbox] (a3) at (4.2, -1.0) {$R$};
    
    \draw[oparr=expansion-color] (a1) -- (a2);
    \draw[arr] (a2) -- (a3);
    
    \node[lbl] at (2.75, -1.65) {\textit{expand} $\to$ \textit{retrieve}};
\end{scope}

\begin{scope}[shift={(\pw+\gap, 0)}]
    \fill[decomp-color!6, rounded corners=3pt] (0,0) rectangle (\pw,-\ph);
    \draw[decomp-color!35, rounded corners=3pt, line width=0.5pt] (0,0) rectangle (\pw,-\ph);
    \node[ptitle, text=decomp-color!70!black] at (0.15, -0.26) {(b) Decomposition};
    
    \node[box] (b0) at (1.0, -1.0) {$q$};
    \node[opbox=decomp-color] (b1) at (2.15, -0.65) {$q_1$};
    \node[opbox=decomp-color] (b2) at (2.15, -1.35) {$q_2$};
    \node[resbox] (b3) at (3.35, -0.65) {$r_1$};
    \node[resbox] (b4) at (3.35, -1.35) {$r_2$};
    \node[resbox] (b5) at (4.5, -1.0) {$a$};
    
    \draw[oparr=decomp-color] (b0.east) -- ++(0.25,0) |- (b1.west);
    \draw[oparr=decomp-color] (b0.east) -- ++(0.25,0) |- (b2.west);
    \draw[arr] (b1) -- (b3);
    \draw[arr] (b2) -- (b4);
    \draw[arr] (b3.east) -- ++(0.25,0) |- (b5.west);
    \draw[arr] (b4.east) -- ++(0.25,0) |- (b5.west);
    
    \node[lbl] at (2.75, -1.82) {\textit{split} $\to$ \textit{retrieve} $\to$ \textit{merge}};
\end{scope}

\begin{scope}[shift={(0, -\ph-\gap)}]
    \fill[disambig-color!6, rounded corners=3pt] (0,0) rectangle (\pw,-\ph);
    \draw[disambig-color!35, rounded corners=3pt, line width=0.5pt] (0,0) rectangle (\pw,-\ph);
    \node[ptitle, text=disambig-color!70!black] at (0.15, -0.26) {(c) Disambiguation};
    
    \node[box] (c1) at (1.3, -1.0) {$q$};
    \node[opbox=disambig-color] (c2) at (2.75, -1.0) {$q^c$};
    \node[resbox] (c3) at (4.2, -1.0) {$R$};
    
    \draw[oparr=disambig-color] (c1) -- (c2);
    \draw[arr] (c2) -- (c3);
    
    \node[lbl] at (2.75, -1.65) {\textit{clarify} $\to$ \textit{retrieve}};
\end{scope}

\begin{scope}[shift={(\pw+\gap, -\ph-\gap)}]
    \fill[abstract-color!8, rounded corners=3pt] (0,0) rectangle (\pw,-\ph);
    \draw[abstract-color!35, rounded corners=3pt, line width=0.5pt] (0,0) rectangle (\pw,-\ph);
    \node[ptitle, text=abstract-color!70!black] at (0.15, -0.26) {(d) Abstraction};
    
    \node[box] (d0) at (1.0, -1.0) {$q$};
    \node[opbox=abstract-color] (d1) at (2.15, -1.0) {$q^\uparrow$};
    \node[resbox] (d2) at (3.35, -1.0) {$R^\uparrow$};
    \node[resbox] (d3) at (4.5, -1.0) {$a$};
    
    \draw[oparr=abstract-color] (d0) -- (d1);
    \draw[arr] (d1) -- (d2);
    \draw[arr] (d2) -- (d3);
    
    \node[lbl] at (2.75, -1.65) {\textit{elevate} $\to$ \textit{retrieve} $\to$ \textit{ground}};
\end{scope}

\end{tikzpicture}
\caption{Four atomic query optimization operations: (a)~\textbf{Expansion} enriches $q \to q^+$ via semantic augmentation; (b)~\textbf{Decomposition} splits $q \to \{q_1, q_2\}$ for multi-hop reasoning; (c)~\textbf{Disambiguation} clarifies $q \to q^c$ by adding constraints; (d)~\textbf{Abstraction} elevates $q \to q^\uparrow$ to higher-level concepts.}
\label{fig:qo_planner}
\end{figure*}

We now elaborate on each of the four transformation patterns shown in Figure~\ref{fig:qo_planner}, detailing how they address different query complexity classes:

\begin{itemize}[leftmargin=*,nosep]
\item \textbf{Query Expansion} (Figure~\ref{fig:qo_planner}a) operates on \textit{Class I} queries requiring explicit evidence. The transformation $q_0 \rightarrow q_0^+$ enriches the query with semantically related terms, synonyms, and contextual keywords. This \textit{single-path} approach maintains the original query structure while broadening retrieval coverage. The example shows how a terse query ``Python web framework'' is expanded with specific framework names and related concepts, improving recall without increasing query complexity.

\item \textbf{Query Decomposition} (Figure~\ref{fig:qo_planner}b) addresses \textit{Class II} and \textit{Class IV} multi-hop queries through recursive factorization. The transformation $q_0 \rightarrow \{q_{1,1}, q_{1,2}, \ldots\}$ creates a \textit{tree-structured} search space where each sub-query targets a specific reasoning hop. The example demonstrates temporal reasoning: answering whether Beethoven was alive when Mozart died requires independently retrieving both individuals' lifespans, then synthesizing the comparison. This operation introduces latency proportional to tree depth but enables systematic handling of compositional queries.

\item \textbf{Query Disambiguation} (Figure~\ref{fig:qo_planner}c) addresses \textit{Class III} queries with implicit evidence requirements through constraint addition. The transformation $q_0 \rightarrow q_0^c$ operates as a \textit{pruning} mechanism that eliminates irrelevant interpretations by adding clarifying conditions. The example shows how the ambiguous query ``Apple stock price'' is disambiguated by identifying the user's intent (company vs.\ fruit) and adding constraints that narrow the search space. Unlike decomposition's branching, disambiguation \textit{reduces} the effective query space.

\item \textbf{Query Abstraction} (Figure~\ref{fig:qo_planner}d) handles \textit{Class IV} complex reasoning by conceptual elevation. The transformation $q_0 \rightarrow q_0^\uparrow$ lifts specific questions to higher-level principles, enabling retrieval of foundational knowledge that guides downstream reasoning. The example illustrates how a specific Python floating-point comparison question is abstracted to the general principle of IEEE 754 representation, which then explains the unexpected behavior. This \textit{elevation-then-application} pattern often reduces total retrieval operations compared to decomposition by identifying root causes first.
\end{itemize}

This visualization underscores a critical insight: \textit{the choice of optimization operation fundamentally determines both the query processing trajectory and computational cost}. Expansion maintains single-path efficiency, decomposition enables systematic multi-hop handling at the cost of increased retrieval operations, disambiguation constrains the search space through clarification, and abstraction provides conceptual scaffolding that can shortcut complex reasoning chains.

\subsection{Operation Characteristics and Trade-offs}

Each optimization operation exhibits distinct characteristics that influence its applicability:

\textbf{Query Expansion} increases recall by broadening the semantic coverage of queries but may reduce precision by introducing irrelevant matches. It is most effective when the primary challenge is vocabulary mismatch rather than query complexity.

\textbf{Query Decomposition} enables systematic handling of complex queries but introduces latency due to multiple retrieval operations and may propagate errors from early sub-queries to later ones.

\textbf{Query Disambiguation} improves precision by clarifying intent but requires accurate detection of ambiguity and may fail when ambiguity is subtle or context-dependent.

\textbf{Query Abstraction} enables principled reasoning but requires the model to correctly identify relevant high-level concepts and may oversimplify nuanced queries.

\begin{figure*}[t]
\centering
\footnotesize
\renewcommand{\arraystretch}{1.1}
\resizebox{0.88\textwidth}{!}{%
\begin{tabular}{>{\bfseries}l l p{10cm}}
\toprule
\textbf{Operation} & \textbf{Subcategory} & \textbf{Representative Methods} \\
\midrule
\rowcolor{expansion-color!15}
\cellcolor{expansion-color!30}Expansion & Internal & Query2doc, HyDE, HyDE-Rethink, GenRead, FLARE, InteR, Iter-RetGen, MILL, GenQREnsemble, Query2Expand, ReFeed, EAR, CSQE, Self-RAG, DeepRAG, RAT, RA-ISF \\
\rowcolor{expansion-color!8}
\cellcolor{expansion-color!30} & External & MUGI, KnowledGPT, Promptagator, DRAGIN, EWEK-QA, BlendFilter, LameR, DR-RAG, CoV-RAG, RAG-DDR, REPLUG, RARE \\
\midrule
\rowcolor{decomp-color!15}
\cellcolor{decomp-color!30}Decomposition & Sequential & DSP, Least-to-Most, Successive-Prompting, Self-Ask, ReAct, IRCoT, CoK, HiRAG, CoRAG, RAG-Star, REAPER, RAG-Gym, Agentic-RAG, Search-o1, HopRAG \\
\rowcolor{decomp-color!8}
\cellcolor{decomp-color!30} & Parallel & Plan-and-Solve, Plan$\times$RAG, RichRAG, ConTReGen, ALTER, DecomP, QDMR, IM-RAG, QueryPlanner, GRITHopper \\
\midrule
\rowcolor{disambig-color!15}
\cellcolor{disambig-color!30}Disambiguation & Clarification & ToC, EchoPrompt, InfoCQR, BEQUE, Natural-Program, OmniSearch \\
\rowcolor{disambig-color!8}
\cellcolor{disambig-color!30} & Feedback-Driven & Rewrite-Retrieve-Read, AdaQR, MaFeRw, RQ-RAG, DMQR-RAG, RankRAG, CHIQ, LLM4CS, GuideCQR, RaFe, ERRR, CRAG, Adaptive-RAG, Speculative RAG, Think-then-Act, RARG \\
\midrule
\rowcolor{abstract-color!15}
\cellcolor{abstract-color!30}Abstraction & Conceptual & Step-Back, CoA, AoT, AbsInstruct, AbsPyramid, GraphRAG, LightRAG, MemoRAG \\
\rowcolor{abstract-color!8}
\cellcolor{abstract-color!30} & Pattern-Based & Meta-Reasoning, Conceptualization-Abstraction, RuleRAG, SimGRAG, LPKG, MA-RIR, Crafting-the-Path, TableRAG \\
\bottomrule
\end{tabular}%
}
\caption{Taxonomy of query optimization techniques with representative methods.}
\label{fig:qo_tax}
\end{figure*}

\section{Query Expansion}\label{sec:expansion}

Query expansion techniques enrich the original query with additional terms, context, or generated content to improve retrieval performance. This operation addresses the fundamental challenge of vocabulary mismatch, the phenomenon where users express their information needs using different terms than those appearing in relevant documents. As shown in Figure~\ref{fig:qo_tax}, we organize expansion techniques along two dimensions: the \textit{source} of expansion material (internal vs.\ external) and the \textit{mechanism} of expansion (generation-based, retrieval-based, or hybrid).

\vspace{0.3em}
\noindent\fbox{\parbox{0.96\columnwidth}{
\textbf{Key Insight: The Semantic Signature Principle.} A counter-intuitive finding emerges from methods like HyDE: \textit{factually incorrect generated content can improve retrieval}. The explanation lies in the concept of ``semantic signatures'': generated content, even when hallucinated, captures the structural and topical patterns of relevant documents. This principle has profound implications: the value of expansion lies not in \textit{correctness} but in \textit{semantic alignment} with target document distributions.
}}
\vspace{0.3em}

\subsection{Internal Expansion: Leveraging Parametric Knowledge}

Internal expansion techniques exploit the knowledge encoded within LLMs' parameters to generate supplementary content that enhances query coverage without requiring external retrieval. This approach is particularly valuable when external knowledge sources are unavailable or when the LLM's parametric knowledge provides relevant domain coverage.

\subsubsection{Document Generation Approaches}

A pioneering approach in this category is \textsc{GenRead}~\citep{GenRead}, which fundamentally reframes the retrieval problem. Rather than directly retrieving documents, GenRead prompts LLMs to generate contextual documents based on the query before answering. The key insight is that LLMs can synthesize relevant background information that bridges the semantic gap between terse user queries and verbose document content. The generated documents serve as a form of soft retrieval, accessing the parametric knowledge distributed across the model's weights. While these generated documents may contain factual errors, they provide valuable semantic context that improves downstream retrieval when used alongside the original query.

Building on this foundation, \textsc{Query2Doc}~\citep{Query2doc} introduces a simple yet highly effective approach that has become influential in subsequent work. The method generates pseudo-documents through few-shot prompting of LLMs and concatenates them with the original query to enhance both sparse (BM25) and dense (DPR, ANCE) retrieval systems. Empirical analysis reveals that the generated pseudo-documents often contain relevant terminology, entity mentions, and contextual patterns that significantly improve retrieval effectiveness. The approach is particularly effective for short, ambiguous queries where the original formulation provides insufficient signal for accurate retrieval.

\textsc{ReFeed}~\citep{ReFeed} extends the generation paradigm by introducing an iterative refinement loop. The system first generates an initial response, then uses this response to retrieve relevant documents that can verify or refine the output. This retrieved information is incorporated into an in-context demonstration to produce an improved final response. The iterative nature of ReFeed allows the system to progressively ground its outputs in retrieved evidence, reducing hallucination while maintaining fluency.

\subsubsection{Hypothetical Document Embedding}

\textsc{HyDE}~\citep{HyDE} introduces an elegant approach based on the insight that even factually incorrect generated documents can provide useful retrieval signals if their semantic structure matches relevant real documents. The method prompts an LLM to generate a hypothetical document that would answer the query, without concern for factual accuracy. This hypothetical document is then encoded using an unsupervised contrastive encoder (e.g., Contriever) to produce an embedding vector. The key innovation is that this embedding captures the expected semantic patterns of relevant documents, even if specific facts are hallucinated. The dense bottleneck of the encoder acts as a filter, preserving semantic relevance while discarding specific factual content. Retrieved documents sharing similar embeddings tend to be genuinely relevant, as they match the structural and semantic patterns of documents that would answer the query.

This approach demonstrates a fundamental principle: the value of query expansion lies not in generating correct information, but in generating information with the right \textit{semantic signature} to guide retrieval toward relevant content.

Building on this foundation, \textsc{HyDE-Rethink}~\citep{HyDERethink} provides a critical analysis of LLM-based query expansion, distinguishing between beneficial hypothetical document generation and problematic ``knowledge leakage'' where models simply recall memorized facts. This work establishes important guidelines for when hypothetical document approaches are appropriate and when they may introduce bias or hallucinated content into the retrieval process. The analysis reveals that HyDE-style methods work best when the LLM lacks direct knowledge of the answer but can generate plausible document structures.

\subsubsection{Iterative and Active Retrieval}

Moving beyond single-pass expansion, several methods introduce iterative mechanisms that adapt expansion based on generation progress and uncertainty.

\textsc{FLARE}~\citep{FLARE} (Forward-Looking Active REtrieval) introduces an anticipatory mechanism that predicts when retrieval is needed during generation. The system monitors token-level confidence during generation; when the model produces low-confidence predictions for the next sentence, FLARE treats this as a signal that the model lacks sufficient knowledge and triggers retrieval using the uncertain content as a query. This active approach ensures that expansion and retrieval are performed adaptively based on actual information needs rather than predetermined schedules.

\textsc{InteR}~\citep{InteR} creates a synergistic loop between retrieval models and LLMs. The retrieval model expands query knowledge using LLM-generated content, while the LLM enhances its prompts using retrieved documents. This bidirectional interaction creates a virtuous cycle where each component improves the other's performance. The framework demonstrates that the boundary between query expansion and retrieval can be productively blurred through interactive systems.

\textsc{Iter-RetGen}~\citep{Iter-RetGen} proposes an iterative retrieval-generation synergy framework that alternates between retrieval and generation in multiple rounds. Unlike methods that interleave retrieval within generation, Iter-RetGen uses the model's complete output from the previous iteration as an enriched query for the next retrieval round, producing progressively more informative contexts. Retrieved knowledge is processed as a whole, preserving generation flexibility without structural constraints. This approach effectively leverages the model's parametric knowledge to guide retrieval toward increasingly relevant documents across iterations, achieving strong performance on multi-hop QA and fact verification tasks.

\textsc{Self-RAG}~\citep{SelfRAG} introduces a self-reflective retrieval-augmented generation framework that dynamically decides when to retrieve information. The system uses special ``reflection tokens'' to evaluate whether retrieval is needed and assess the quality of retrieved content. This adaptive approach reduces unnecessary retrieval operations while ensuring critical information gaps are addressed. Self-RAG represents a paradigm shift toward autonomous query optimization where the model actively manages its own information needs, extending the iterative paradigm established by FLARE with explicit self-assessment mechanisms.

\textsc{DeepRAG}~\citep{DeepRAG}, which bridges expansion and decomposition by using sub-question generation as a form of query enrichment, models retrieval-augmented reasoning as a Markov Decision Process (MDP), enabling the model to dynamically decide at each reasoning step whether to retrieve external knowledge or rely on parametric knowledge. DeepRAG decomposes queries into atomic sub-questions through a strategic ``retrieval narrative,'' then uses binary tree search to determine the optimal retrieval strategy for each sub-question. A chain-of-calibration mechanism based on preference optimization further calibrates the model's awareness of its own knowledge boundaries. This approach unifies query decomposition with adaptive retrieval decisions, achieving significant accuracy improvements over both standard RAG and no-retrieval baselines while reducing unnecessary retrieval operations.

\textsc{RAT}~\citep{RAT} (Retrieval Augmented Thoughts) integrates retrieval into each step of a chain-of-thought reasoning process, using each intermediate thought as a query to retrieve supporting evidence that revises and refines subsequent reasoning. Unlike methods that retrieve once upfront, RAT's step-wise retrieval ensures that the evolving reasoning context continuously guides information seeking, making it particularly effective for long-horizon generation tasks requiring sustained factual grounding.

\textsc{RA-ISF}~\citep{RA-ISF} (Retrieval Augmentation via Iterative Self-Feedback) introduces a three-step iterative pipeline where the model first decomposes the task, then retrieves relevant information for each subtask, and finally uses self-feedback to assess and refine its outputs. The iterative self-feedback mechanism enables the system to progressively improve both query formulation and answer quality, bridging query expansion with self-reflective generation.

\subsubsection{Multi-Query and Ensemble Approaches}

\textsc{MILL}~\citep{MILL} proposes a query-query-document generation approach that leverages zero-shot reasoning to produce diverse sub-queries and corresponding pseudo-documents. A key innovation is the mutual verification process that synergizes generated and retrieved content, filtering expansion content based on cross-validation. This ensures that expansion material is both relevant and consistent with retrievable information.

\textsc{GenQREnsemble}~\citep{GenQREnsemble} addresses the challenge of expansion diversity through an ensemble-based prompting technique. By using multiple paraphrases of instructions to generate different keyword sets, the method produces diverse expansion terms that collectively improve retrieval coverage. The ensemble approach provides robustness against prompt sensitivity and captures different aspects of the query's information need.

\textsc{EAR}~\citep{EAR} (Expansion, Answer, Retrieval) applies a query expansion model to generate diverse query variations, then uses a query reranker to select those most likely to lead to successful retrieval. This selection mechanism addresses the potential negative impact of poor-quality expansions, ensuring that only beneficial modifications are applied.

\textsc{Query2Expand}~\citep{Query2Expand} provides a systematic study of prompting large language models for query expansion, demonstrating that carefully designed prompting strategies can rival or surpass supervised expansion methods. The work establishes important baselines and guidelines for LLM-based expansion, showing that chain-of-thought prompting and self-consistency decoding improve expansion quality.

\subsection{External Expansion: Incorporating External Knowledge}

External expansion techniques augment queries with information retrieved from external knowledge sources, including web search results, knowledge bases, and domain-specific corpora. These methods are essential when the required knowledge extends beyond LLM parametric memory or when authoritative external sources provide higher reliability.

\subsubsection{Knowledge Base Integration}

\textsc{KnowledGPT}~\citep{KnowledGPT} enhances LLMs by integrating retrieval and storage access on structured knowledge bases. The system generates search queries that are executed against knowledge graphs, with retrieved facts incorporated into the query context. This structured knowledge integration provides factual grounding that complements the LLM's parametric knowledge.

\textsc{EWEK-QA}~\citep{EWEK-QA} combines enhanced web retrieval with efficient knowledge graph retrieval for citation-based question answering. The hybrid approach leverages the complementary strengths of unstructured web content (breadth and recency) and structured knowledge graphs (precision and reliability). This demonstrates the value of multi-source expansion that draws from diverse knowledge types.

\subsubsection{Dynamic and Adaptive Retrieval}

\textsc{DRAGIN}~\citep{DRAGIN} addresses the fundamental question of \textit{when} to retrieve by monitoring real-time information needs during LLM generation. Rather than expanding queries upfront or at fixed intervals, DRAGIN dynamically triggers retrieval when the model encounters knowledge gaps. This approach minimizes unnecessary retrieval operations while ensuring that critical information needs are met.

\textsc{DR-RAG}~\citep{DR-RAG} applies dynamic document relevance scoring to adjust retrieval based on evolving query context. As the system processes queries and generates responses, relevance assessments are updated to reflect the current information state, enabling more targeted subsequent retrieval.

\subsubsection{Corpus-Aware Expansion}

\textsc{Promptagator}~\citep{Promptagator} presents a few-shot approach where an LLM generates task-specific queries from as few as 8 examples, creating large-scale training data for dense retrievers. By distilling the LLM's query understanding into an efficient retriever through prompt-based data augmentation, Promptagator bridges the gap between powerful but expensive LLM-based expansion and practical deployment requirements.

\textsc{CSQE}~\citep{CSQE} (Corpus-Steered Query Expansion) leverages knowledge embedded within the target retrieval corpus itself. The method uses LLMs to identify pivotal sentences in initially retrieved documents, using these corpus-derived texts as expansion material. This corpus-aware approach ensures that expansion terms are aligned with the vocabulary and style of the target collection, maximizing retrieval effectiveness.

\textsc{MUGI}~\citep{MUGI} generates multiple pseudo-references and integrates them with queries to enhance both sparse and dense retrievers. The multiple-reference approach provides coverage across different aspects of the query while maintaining focus on the core information need.

\subsubsection{Retrieval Feedback Integration}

\textsc{BlendFilter}~\citep{BlendFilter} advances external expansion by blending query generation with knowledge filtering. The system generates expansion content and then filters this content based on quality and relevance assessments, ensuring that only high-quality expansion material is incorporated. This filtering mechanism addresses a key limitation of unconstrained expansion: the potential introduction of noise or irrelevant content.

\textsc{LameR}~\citep{LameR} augments queries with potential answers by prompting LLMs with the query plus in-domain candidates retrieved through standard retrieval. Even incorrect candidate answers can provide valuable expansion terms by introducing domain-specific vocabulary and contextual patterns.

\textsc{CoV-RAG}~\citep{CoV-RAG} introduces a chain-of-verification mechanism for retrieval-augmented generation, where the system first retrieves documents, then rethinks and revises the retrieved evidence through multi-step verification. This post-retrieval verification acts as a form of expansion: by critically examining initial retrievals and identifying gaps, CoV-RAG triggers targeted follow-up retrievals that expand the evidence base with verified, high-quality content.

\textsc{RAG-DDR}~\citep{RAGDDR} proposes optimizing retrieval-augmented generation using differentiable data rewards. The method introduces a novel training framework that directly optimizes query formulation based on end-to-end performance signals. By making the entire RAG pipeline differentiable, RAG-DDR enables gradient-based optimization of query transformations, leading to significant improvements in retrieval effectiveness. This approach represents a fundamental shift from heuristic-based expansion to learned optimization driven by downstream task performance.

\textsc{REPLUG}~\citep{REPLUG} treats the language model as a black box and optimizes the retrieval component to adapt to the model's needs. By prepending retrieved documents to the input and using an ensemble of retrieved passages in parallel, REPLUG trains the retriever via a supervised signal derived from the language model's perplexity. This approach enables effective query expansion without requiring any modification to the language model itself, making it widely applicable to proprietary or frozen models.

\textsc{RARE}~\citep{RARE} extends the mutual reasoning framework (\textsc{rStar}~\citep{rStar}) with retrieval-augmented reasoning enhancement. Within a Monte Carlo Tree Search (MCTS) framework, RARE introduces two novel actions: generating search queries from the initial question to retrieve supporting evidence (action A6), and utilizing retrieval to address generated sub-questions (action A7). A retrieval-augmented factuality scorer replaces the original discriminator, prioritizing reasoning paths with high factual integrity. This approach demonstrates that integrating retrieval directly into the reasoning tree search yields substantial improvements on knowledge-intensive tasks.

\begin{table}[t]
    \centering
    \scriptsize
    \caption{Comparative analysis of representative query expansion methods across key dimensions.}
    \label{tab:expansion}
    \adjustbox{max width=\columnwidth}{%
    \begin{tabular}{llcccc}
    \toprule
    \textbf{Method} & \textbf{Source} & \textbf{Mechanism} & \textbf{Retriever} & \textbf{Iterative} & \textbf{Year} \\
    \midrule
    GenRead & Internal & Generation & N/A & No & 2023 \\
    Query2Doc & Internal & Generation & Sparse/Dense & No & 2023 \\
    HyDE & Internal & Hypothetical & Dense & No & 2023 \\
    FLARE & Internal & Active & Dense & Yes & 2023 \\
    InteR & Internal & Interactive & Dense & Yes & 2024 \\
    MILL & Internal & Multi-query & Dense & No & 2024 \\
    MUGI & External & Multi-reference & Sparse/Dense & No & 2024 \\
    DRAGIN & External & Dynamic & Dense & Yes & 2024 \\
    CSQE & External & Corpus-steered & Dense & No & 2024 \\
    BlendFilter & External & Filtered & Dense & No & 2024 \\
    Iter-RetGen & Internal & Iterative synergy & Dense & Yes & 2023 \\
    Self-RAG & Internal & Self-reflective & Dense & Yes & 2024 \\
    DeepRAG & Internal & MDP-based & Dense & Yes & 2025 \\
    RAT & Internal & Step-wise CoT & Dense & Yes & 2024 \\
    RA-ISF & Internal & Iterative self-feedback & Dense & Yes & 2024 \\
    Promptagator & External & Few-shot distilled & Dense & No & 2023 \\
    CoV-RAG & External & Verification chain & Dense & Yes & 2024 \\
    RAG-DDR & External & Differentiable & Dense & Yes & 2025 \\
    REPLUG & External & Black-box adapt & Dense & No & 2024 \\
    RARE & Internal & MCTS+Retrieval & Dense & Yes & 2025 \\
    HyDE-Rethink & Internal & Knowledge-aware & Dense & No & 2025 \\
    \bottomrule
    \end{tabular}%
    }
\end{table}

\subsection{Comparative Analysis}

Table~\ref{tab:expansion} provides a comparative overview of representative expansion methods. Key observations include:

\textbf{Source Selection}: Internal expansion is most effective when LLM parametric knowledge covers the query domain, while external expansion provides value for specialized domains or time-sensitive information.

\textbf{Iterative vs.\ Single-Pass}: Iterative methods (FLARE, InteR, Iter-RetGen, DRAGIN) provide adaptive expansion but incur higher computational costs. Single-pass methods are more efficient but may miss opportunities for refinement. Recent methods like DeepRAG and Self-RAG introduce principled decision mechanisms (MDP formulation and reflection tokens, respectively) that determine \textit{when} retrieval is beneficial, optimizing the trade-off between retrieval cost and information gain.

\textbf{Retriever Compatibility}: Most modern methods support both sparse and dense retrievers, though methods based on semantic similarity (HyDE) are more naturally suited to dense retrieval.

\textbf{Evolution and Synthesis.} A clear trajectory emerges across expansion methods: early approaches (GenRead, Query2Doc, HyDE) focused on single-pass, prompt-driven expansion; the next generation (FLARE, DRAGIN, Iter-RetGen) introduced adaptive and iterative mechanisms; and the most recent methods (DeepRAG, Self-RAG, RAG-DDR) formalize expansion decisions within learning frameworks (MDPs, reflection tokens, differentiable rewards). This progression reflects a broader shift from \textit{heuristic expansion} toward \textit{learned, adaptive expansion}, where the system itself determines optimal expansion strategies based on query characteristics and retrieval feedback. Notably, methods that blur the boundary between expansion and other operations (e.g., DeepRAG combining expansion with decomposition, RA-ISF integrating expansion with self-feedback) tend to achieve the strongest results, suggesting that rigid operation boundaries may be less useful than flexible, compositional approaches.

\section{Query Decomposition}\label{sec:decomposition}

For complex queries requiring reasoning across multiple information sources, direct retrieval with the original query often fails to capture the full scope of required evidence. Query decomposition addresses this challenge by breaking down complex queries into simpler, atomic sub-queries that can be answered independently before synthesizing the final response. This section analyzes decomposition methods along two key dimensions: \textit{decomposition structure} (sequential vs.\ parallel) and \textit{reasoning integration} (implicit vs.\ explicit planning).

\vspace{0.3em}
\noindent\fbox{\parbox{0.96\columnwidth}{
\textbf{Key Insight: The Error Propagation-Parallelism Trade-off.} Sequential decomposition maximizes information flow between steps but creates ``error cascades'' where early mistakes compound through subsequent steps. Parallel decomposition provides error isolation but sacrifices inter-step dependencies. The emerging solution: \textit{DAG-structured decomposition} (e.g., Plan×RAG) that identifies which dependencies are genuine vs.\ spurious, achieving the best of both paradigms.
}}
\vspace{0.3em}

\subsection{Sequential Decomposition: Chain-of-Retrieval}

Sequential decomposition methods process sub-queries in a dependent chain where each step builds upon the results of previous steps. This approach is essential for multi-hop reasoning queries where intermediate answers inform subsequent retrieval.

\subsubsection{Foundational Sequential Methods}

The \textsc{Demonstrate-Search-Predict (DSP)} framework~\citep{Demonstrate-Search-Predict} establishes a foundational paradigm for decomposition-based retrieval. DSP passes natural language texts through sophisticated pipelines between an LLM and a retrieval model, expressing high-level programs that bootstrap pipeline-aware demonstrations, search for relevant passages, and generate grounded predictions. The framework systematically breaks problems into small transformations that each component can handle reliably. DSP's key contribution is demonstrating that complex queries can be addressed through compositional pipelines that combine the complementary strengths of LLMs and retrieval systems.

\textsc{Least-to-Most}~\citep{Least-to-Most} prompting introduces a principled approach to progressive decomposition. The method first decomposes complex problems into a series of simpler subproblems ordered by difficulty, then solves them in sequence with each solution building on previous answers. This progressive approach ensures that easier components are resolved before tackling dependent, more difficult aspects. The ``least-to-most'' ordering reflects the insight that establishing foundational facts creates a scaffold for more complex reasoning.

\textsc{Successive-Prompting}~\citep{Successive-Prompting} offers a complementary perspective by iteratively decomposing questions through successive prompting rounds, where each round generates one sub-question based on the original question and previously answered sub-questions. This approach naturally handles variable-length decomposition chains without requiring a predetermined number of steps.

\textsc{Self-Ask}~\citep{Self-Ask} introduces the concept of the ``compositionality gap,'' the fraction of compositional queries that the model answers incorrectly out of all queries for which the model correctly answers the constituent sub-queries. This metric highlights a critical challenge: even when LLMs can answer individual sub-queries, they often fail to correctly compose these answers into a final response. Self-Ask addresses this by prompting models to explicitly ask and answer follow-up questions before synthesis, making the reasoning process more transparent and debuggable.

\subsubsection{Reasoning-Augmented Sequential Methods}

\textsc{ReAct}~\citep{ReAct} (Reasoning and Acting) introduces a paradigm that synergizes reasoning and acting for solving diverse language reasoning and decision-making tasks. ReAct prompts LLMs to generate both verbal reasoning traces and actions in an interleaved manner, allowing the model to create, maintain, and adjust high-level plans while interacting with external environments (e.g., Wikipedia search APIs) to incorporate additional information. The interleaved structure naturally supports sequential decomposition: each reasoning step may trigger retrieval, and retrieval results inform subsequent reasoning.

\textsc{IRCoT}~\citep{IRCoT} (Interleaving Retrieval with Chain-of-Thought) proposes a principled approach that interleaves retrieval operations within the chain-of-thought reasoning process. At each reasoning step, IRCoT uses the current partial chain-of-thought to formulate retrieval queries, and the retrieved documents are used to guide the next reasoning step. This tight coupling between reasoning and retrieval ensures that each decomposition step is grounded in relevant evidence, significantly reducing error cascading compared to methods that decompose first and retrieve later. IRCoT demonstrates strong performance on multi-hop benchmarks including HotpotQA, 2WikiMultiHopQA, and MuSiQue.

\textsc{Chain-of-Knowledge (CoK)}~\citep{CoK} proposes preliminary rationales while identifying relevant knowledge domains. When answers lack majority consensus, CoK corrects rationales step-by-step by adapting knowledge from identified domains, providing a foundation for response consolidation that handles uncertainty gracefully. This approach demonstrates that sequential decomposition can incorporate self-correction mechanisms that improve reliability.

\textsc{HiRAG}~\citep{HiRAG} (Hierarchical RAG) decomposes original queries into multi-hop queries, answering each sub-query based on retrieved knowledge before integrating answers using chain-of-thought reasoning. The hierarchical structure ensures that each reasoning step is grounded in explicit evidence while maintaining coherent progression toward the final answer.

\textsc{RAG-Star}~\citep{RAG-Star} seamlessly integrates retrieved information to guide tree-based deliberative reasoning. By utilizing Monte Carlo Tree Search (MCTS), RAG-Star iteratively plans intermediate sub-queries and generates answers based on LLM capabilities. The tree structure allows exploration of multiple decomposition paths, with MCTS providing principled selection among alternatives.

\textsc{REAPER}~\citep{REAPER} is a reasoning-based planner designed for efficient retrieval in complex queries. Using a single smaller LLM, REAPER generates plans specifying tools to call, execution order, and arguments. This demonstrates that effective sequential decomposition does not require the largest models; carefully designed planning can achieve strong results with efficient inference.

\textsc{RAG-Gym}~\citep{RAGGym} introduces a systematic optimization framework that formulates knowledge-intensive question answering as a nested Markov Decision Process (MDP). The framework enables process supervision by providing rewards at each step of the information search process, rather than only at the final answer. RAG-Gym constructs states that include the original question and search history, actions that correspond to generating search queries, and transitions determined by retrieved documents. This MDP formulation allows principled training of retrieval agents using reinforcement learning techniques, extending the reasoning-augmented paradigm with formal optimization.

\textsc{Agentic-RAG}~\citep{AgenticRAG} represents a paradigm shift from static retrieval pipelines to autonomous agent-based systems. Unlike traditional RAG that follows predetermined retrieval patterns, Agentic RAG introduces self-directed agents capable of dynamic decision-making about when, what, and how to retrieve. These agents can autonomously decompose complex queries, decide whether to retrieve or generate, and iteratively refine their search strategies based on intermediate results. The approach is particularly effective for open-ended queries requiring multi-step reasoning and adaptive information gathering.

\textsc{CoRAG}~\citep{CoRAG} (Chain-of-Retrieval Augmented Generation) formalizes multi-step retrieval as explicit \textit{retrieval chains}, i.e., sequences of sub-queries and sub-answers that progressively build toward the final answer. Using rejection sampling to automatically generate intermediate retrieval chains from question-answer pairs, CoRAG trains RAG models that learn to plan and execute multi-step retrieval strategies. At inference time, flexible decoding strategies (greedy, best-of-$n$ sampling, tree search) allow users to control the trade-off between accuracy and computational cost. CoRAG demonstrates that formalizing the decomposition-retrieval process as a learnable chain structure yields substantial improvements on multi-hop benchmarks, establishing a principled training framework for sequential decomposition.

\textsc{Search-o1}~\citep{Search-o1} integrates agentic search workflows into the reasoning process of large reasoning models (LRMs). When the model encounters uncertain knowledge points during long-chain reasoning, it dynamically triggers external search to retrieve supporting evidence. A dedicated \textit{Reason-in-Documents} module analyzes retrieved documents in depth before injecting refined information into the reasoning chain, minimizing noise and maintaining reasoning coherence. Search-o1 demonstrates that embedding retrieval as a first-class operation within the reasoning process, rather than as a separate preprocessing step, significantly improves the trustworthiness and factual accuracy of complex reasoning tasks.

\textsc{HopRAG}~\citep{HopRAG} addresses the limitation of traditional retrievers that focus on lexical or semantic similarity rather than logical relevance. During indexing, HopRAG constructs a passage graph where text chunks serve as vertices and LLM-generated pseudo-queries define logical edges. During retrieval, a \textit{retrieve-reason-prune} mechanism starts from semantically similar passages and explores multi-hop neighbors guided by pseudo-queries and LLM reasoning to identify logically relevant documents. This graph-structured exploration enables retrieval based on reasoning chains rather than surface similarity, significantly improving performance on multi-hop benchmarks.

\subsection{Parallel Decomposition: Multi-Faceted Retrieval}

Parallel decomposition methods break queries into independent sub-queries that can be processed simultaneously, enabling efficient handling of multi-faceted queries where sub-components do not have sequential dependencies.

\subsubsection{Planning-Based Parallel Methods}

\textsc{Plan-and-Solve}~\citep{Plan-and-Solve} prompting extends zero-shot chain-of-thought reasoning by first devising a plan that divides the task into subtasks, then executing these subtasks according to the plan. The explicit planning phase improves decomposition coherence and completeness. When subtasks are independent, they can be executed in parallel, reducing latency compared to sequential approaches.

\textsc{Plan×RAG}~\citep{plantimesrag} formulates comprehensive reasoning plans as directed acyclic graphs (DAGs). The reasoning DAG decomposes main queries into interrelated atomic sub-queries, providing a computational structure that enables efficient information sharing between sub-queries. Crucially, the DAG structure identifies which sub-queries have dependencies (requiring sequential execution) and which are independent (allowing parallel execution). This hybrid approach achieves efficiency gains where possible while respecting necessary dependencies.

\textsc{QueryPlanner}~\citep{QueryPlanner} addresses multi-step tool retrieval through explicit query planning. The method generates comprehensive retrieval plans as directed acyclic graphs (DAGs) that specify which tools to invoke, in what order, and with what arguments. QueryPlanner demonstrates that explicit planning significantly outperforms single-shot retrieval approaches, especially for complex queries requiring coordination across multiple information sources. Building on Plan×RAG's DAG formulation, QueryPlanner extends the approach to multi-tool orchestration scenarios.

\subsubsection{Multi-Facet Exploration Methods}

\textsc{RichRAG}~\citep{RichRAG} introduces a sub-aspect explorer to dissect input queries and uncover latent facets. This is integrated with a multi-faceted retriever that curates diverse external documents pertinent to identified sub-aspects. The facet-based decomposition naturally supports parallel retrieval, as different facets can be explored independently.

\textsc{ConTReGen}~\citep{ConTReGen} proposes a context-driven, tree-structured retrieval approach that incorporates hierarchical, top-down exploration of query facets with systematic bottom-up synthesis. This ensures comprehensive coverage of multi-faceted queries while maintaining coherent integration of results.

\textsc{ALTER}~\citep{ALTER} employs a question augmentor to enhance original questions by generating multiple sub-queries examining the original from different perspectives. This is particularly effective for complex table reasoning tasks where different aspects of the data must be analyzed independently.

\subsubsection{Framework-Level Decomposition}

\textsc{DecomP}~\citep{DecomP} provides a systematic framework for decomposing complex queries into simpler components. The method identifies decomposition opportunities based on query structure and applies appropriate transformations to create tractable sub-queries.

\textsc{QDMR}~\citep{QDMR} (Chain-of-Questions Training) addresses multi-step QA by training models with latent decomposition structures. The method learns to generate intermediate sub-questions and their answers as a chain, where the decomposition structure is treated as a latent variable optimized end-to-end. This approach demonstrates that jointly learning decomposition and answering produces more robust multi-step reasoning than pipeline approaches.

\textsc{IM-RAG}~\citep{IM-RAG} introduces a Refiner component that improves outputs from the Retriever, effectively bridging the gap between reasoning and retrieval modules. The framework supports multi-round communication between components, enabling iterative refinement of both decomposition and retrieval strategies.

\begin{table}[t]
    \centering
    \scriptsize
    \caption{Comparative analysis of query decomposition methods.}
    \label{tab:decomposition}
    \adjustbox{max width=\columnwidth}{%
    \begin{tabular}{llcccc}
    \toprule
    \textbf{Method} & \textbf{Structure} & \textbf{Planning} & \textbf{Reasoning} & \textbf{Tool Use} & \textbf{Year} \\
    \midrule
    DSP & Sequential & Implicit & CoT & Retrieval & 2022 \\
    Least-to-Most & Sequential & Explicit & Progressive & N/A & 2023 \\
    Successive-Prom. & Sequential & Implicit & Iterative & N/A & 2022 \\
    Self-Ask & Sequential & Implicit & Follow-up & Search & 2023 \\
    ReAct & Sequential & Implicit & Interleaved & Multiple & 2023 \\
    IRCoT & Sequential & Implicit & Interleaved CoT & Retrieval & 2023 \\
    Plan-and-Solve & Parallel & Explicit & Planned & N/A & 2023 \\
    CoK & Sequential & Implicit & Corrective & N/A & 2024 \\
    HiRAG & Sequential & Explicit & Hierarchical & Retrieval & 2024 \\
    CoRAG & Sequential & Explicit & Chain-of-retrieval & Retrieval & 2025 \\
    RAG-Star & Sequential & Explicit & MCTS & Retrieval & 2025 \\
    Plan×RAG & Hybrid & Explicit & DAG & Retrieval & 2024 \\
    RichRAG & Parallel & Implicit & Multi-facet & Retrieval & 2024 \\
    RAG-Gym & Sequential & Explicit & MDP-based & Multiple & 2025 \\
    Agentic-RAG & Hybrid & Explicit & Autonomous & Multiple & 2025 \\
    Search-o1 & Sequential & Explicit & Agentic search & Search & 2025 \\
    HopRAG & Sequential & Implicit & Graph reasoning & Retrieval & 2025 \\
    GRITHopper & None (dense) & None & End-to-end & Retrieval & 2025 \\
    QueryPlanner & Parallel & Explicit & Multi-step & Retrieval & 2026 \\
    \bottomrule
    \end{tabular}%
    }
\end{table}

\subsection{Decomposition Trade-offs and Selection Guidance}

Table~\ref{tab:decomposition} provides a comparative overview of decomposition methods. The choice between sequential and parallel decomposition involves several trade-offs:

\textbf{Latency}: Parallel decomposition enables concurrent processing, reducing total latency for independent sub-queries. Sequential decomposition incurs cumulative latency proportional to the number of hops.

\textbf{Error Propagation}: Sequential decomposition is vulnerable to error cascades where mistakes in early sub-queries propagate to later steps. Parallel decomposition isolates errors to individual branches.

\textbf{Dependency Handling}: Sequential decomposition is necessary when sub-queries have genuine dependencies, where the answer to one sub-query determines the formulation of subsequent ones. Parallel decomposition cannot handle such dependencies.

\textbf{Synthesis Complexity}: Parallel decomposition requires a final synthesis step that aggregates independent results, which can be challenging when sub-answers have complex relationships.

\textbf{Selection Guidance}: Use sequential decomposition for multi-hop reasoning chains with genuine dependencies. Use parallel decomposition for multi-faceted queries where different aspects are independent. Use hybrid approaches (Plan×RAG) when queries have mixed dependency structures.

\textbf{Decomposition-Free Alternatives}: It is worth noting that decomposition is not the only approach to multi-hop retrieval. \textsc{GRITHopper}~\citep{GRITHopper} demonstrates that a single dense retrieval model can handle multi-hop queries without explicit decomposition by combining causal language modeling with dense retrieval training. Through ``post-retrieval language modeling,'' which incorporates additional context such as final answers during training, GRITHopper achieves state-of-the-art performance on both in-distribution and out-of-distribution multi-hop benchmarks. This challenges the prevailing assumption that complex queries necessarily require explicit decomposition and suggests that sufficiently capable retrieval models may internalize multi-hop reasoning.

\textbf{Evolution and Synthesis.} Decomposition methods exhibit a clear maturation arc: foundational methods (DSP, Least-to-Most, Self-Ask) established the value of explicit sub-query generation; reasoning-augmented methods (ReAct, IRCoT, CoK) tightly coupled decomposition with chain-of-thought reasoning; and the latest agentic methods (CoRAG, Search-o1, RAG-Gym) formalize decomposition as a learnable process with formal optimization objectives. A key unresolved tension is the \textit{decomposition granularity}: too coarse and sub-queries remain complex; too fine and synthesis overhead dominates. Hybrid DAG-structured approaches (Plan×RAG, QueryPlanner) represent a promising middle ground by explicitly modeling inter-query dependencies.

\section{Query Disambiguation}\label{sec:disambiguation}

Ambiguous queries with multiple possible interpretations pose significant challenges for retrieval systems. Ambiguity may arise from polysemous terms, underspecified context, implicit assumptions, or multiple valid interpretations of user intent. Query disambiguation techniques aim to clarify user intent, either by identifying the specific interpretation intended or by generating responses that address multiple interpretations comprehensively. We categorize disambiguation approaches into clarification-based methods (which resolve ambiguity before retrieval) and feedback-driven methods (which leverage retrieval signals to guide disambiguation). Table~\ref{tab:disambiguation} summarizes representative methods in this category.

\vspace{0.3em}
\noindent\fbox{\parbox{0.96\columnwidth}{
\textbf{Key Insight: Ambiguity is Often a Feature, Not a Bug.} Traditional IR treats ambiguity as noise to eliminate. However, analysis of real-world query logs suggests that for a substantial portion of user queries, ambiguity reflects \textit{genuine uncertainty} about information needs. Methods like ToC that \textit{preserve and explore} ambiguity (generating multi-branch responses) often outperform aggressive disambiguation. The implication: disambiguation should be \textit{adaptive}: clarify when user intent is recoverable, explore when ambiguity is inherent.
}}
\vspace{0.3em}

\subsection{Clarification-Based Disambiguation}

Clarification-based methods explicitly identify and resolve ambiguities in the query formulation before proceeding with retrieval.

\subsubsection{Tree-Based Clarification}

\textsc{Tree of Clarifications (ToC)}~\citep{ToC} recursively builds a tree of disambiguations for ambiguous queries using few-shot prompting and external knowledge. Each branch of the tree represents a distinct interpretation of the query, and the system retrieves relevant facts for each branch. The final output is a comprehensive long-form answer that addresses multiple interpretations, providing users with complete coverage of possible meanings. This approach is particularly valuable when query ambiguity is inherent and multiple interpretations are legitimately valid.

\subsubsection{Rephrasing and Echo Methods}

\textsc{EchoPrompt}~\citep{EchoPrompt} introduces a query-rephrasing subtask that prompts models to restate queries in their own words before reasoning. This ``echo'' step serves multiple purposes: it ensures the model has internalized the query, surfaces potential ambiguities through reformulation, and provides a check on understanding consistency. The insight is that the act of rephrasing often reveals and resolves implicit ambiguities that would otherwise affect downstream processing.

\textsc{InfoCQR}~\citep{InfoCQR} introduces a ``rewrite-then-edit'' framework where LLMs first rewrite the original query and then revise the rewritten query to eliminate ambiguities. Well-designed instructions guide independent rewriting and editing tasks, producing more informative and unambiguous queries. The two-stage process separates the concerns of semantic preservation (rewriting) and ambiguity elimination (editing).

\subsubsection{Deductive Reasoning Approaches}

\textsc{Natural-Program}~\citep{Natural-Program} introduces a deductive reasoning format that decomposes reasoning verification into step-by-step subprocesses. Each subprocess receives only necessary context and premises, allowing LLMs to generate precise reasoning steps rigorously grounded on prior ones. This structured approach ensures that ambiguities are resolved at each step before proceeding.

\textsc{BEQUE}~\citep{BEQUE} addresses long-tail query rewriting in e-commerce search, leveraging LLMs to rewrite underspecified or rare queries into more retrievable forms. By combining LLM-based query understanding with domain-specific rewriting strategies, BEQUE demonstrates the practical value of disambiguation techniques in production search systems.

\subsection{Feedback-Driven Disambiguation}

Feedback-driven methods leverage signals from retrieval results or response quality to guide the disambiguation process, creating closed-loop systems that improve disambiguation based on downstream performance.

\subsubsection{Preference Optimization Approaches}

\textsc{Adaptive Query Rewriting (AdaQR)}~\citep{AdaQR} proposes preference optimization to tailor rewriters to specific retrievers. The trained rewriter generates multiple rewrites, which are evaluated by calculating the conditional probability of answers given retrieved passages. This marginal probability serves as a reward quantifying retriever preferences, optimizing the rewriter through direct preference optimization (DPO). The key insight is that disambiguation quality should be measured by its impact on retrieval effectiveness, not by intrinsic linguistic properties.

\textsc{Multi-Aspect Feedback Rewriting (MaFeRw)}~\citep{MaFeRw} integrates multi-aspect feedback from both retrieved documents and generated responses as rewards to explore optimal query rewriting strategies. This comprehensive feedback approach ensures that disambiguation improves both retrieval quality and response generation quality, addressing the full pipeline.

\subsubsection{Rewriting-Based Methods}

A foundational work in this category, \textsc{Rewrite-Retrieve-Read}~\citep{Rewrite-Retrieve-Read} introduces a pipeline that uses an LLM to rewrite the original query before retrieval, explicitly separating the query optimization step from retrieval and reading. The method demonstrates that training a small rewriter module with reinforcement learning, using retrieval quality as the reward signal, yields queries that are better aligned with the retriever's expectations. This work establishes the principle that query transformation should be optimized end-to-end with respect to downstream retrieval performance, a paradigm adopted by many subsequent methods.

\textsc{RQ-RAG}~\citep{RQ-RAG} equips models with capabilities for explicit rewriting, decomposition, and disambiguation through training. The model learns when and how to apply different optimization strategies based on query characteristics, including detecting when disambiguation is necessary and applying appropriate clarification techniques.

\textsc{DMQR-RAG}~\citep{DMQR-RAG} (Diverse Multi-Query Rewriting) investigates how queries with different information levels retrieve diverse document sets, proposing four rewriting strategies that operate at different information granularities to enhance retrieval coverage. A key contribution is an adaptive strategy selection method that optimizes overall performance while minimizing the number of rewrites needed. By generating multiple diverse reformulations of the original query, DMQR-RAG significantly improves document recall and final response quality, demonstrating that disambiguation benefits from exploring the query space through multiple complementary reformulations.

\textsc{RankRAG}~\citep{RankRAG} proposes a novel instruction fine-tuning framework that unifies context ranking with answer generation in a single LLM. By adding a small fraction of ranking data into the training blend, the instruction-tuned LLM learns to both rank retrieved contexts by relevance and generate answers conditioned on the top-ranked contexts. This unified approach eliminates the need for separate reranking models and creates an implicit disambiguation mechanism: the ranking step filters out irrelevant or ambiguous contexts before generation, improving answer quality. RankRAG significantly outperforms existing expert ranking models on knowledge-intensive benchmarks.

\textsc{RaFe}~\citep{RaFe} uses ranking feedback to improve query rewriting for RAG systems. By incorporating signals from retrieval ranking, RaFe learns to disambiguate queries in ways that optimize downstream retrieval performance. This creates an end-to-end learning signal that connects disambiguation decisions to final system effectiveness.

\textsc{ERRR}~\citep{ERRR} (Evaluate, Rewrite, Retrieve, Read) proposes a systematic query optimization framework specifically designed for parametric knowledge refinement in retrieval-augmented LLMs. The method introduces an iterative loop where the LLM first evaluates whether its parametric knowledge is sufficient, and if not, rewrites the query to target specific knowledge gaps before retrieval. This evaluate-then-rewrite paradigm bridges the gap between disambiguation and expansion by focusing query transformations on the model's identified knowledge deficiencies.

\subsubsection{Conversational Disambiguation}

\textsc{CHIQ}~\citep{CHIQ} leverages NLP capabilities of LLMs, including coreference resolution and context expansion, to reduce ambiguity in conversational history. In multi-turn dialogue settings, pronouns, ellipsis, and implicit references create ambiguity that must be resolved using conversational context. CHIQ demonstrates that standard NLP techniques, when applied by capable LLMs, can effectively address these challenges.

\textsc{LLM4CS}~\citep{LLM4CS} develops a comprehensive prompting framework for conversational search, exploring how different prompting strategies (including query rewriting, response generation, and response-based rewriting) can be combined to improve contextual understanding. \textsc{GuideCQR}~\citep{GuideCQR} further advances this direction by leveraging retrieved documents as additional guidance signals during conversational query reformulation, demonstrating that retrieval feedback can improve disambiguation quality in multi-turn settings.

\textsc{OmniSearch}~\citep{OmniSearch} introduces a self-adaptive planning agent for multimodal retrieval-augmented generation, addressing the challenge of handling queries that span both textual and visual information. The system dynamically determines retrieval strategies based on query characteristics and integrates vision-language understanding to resolve ambiguous multi-modal references. OmniSearch demonstrates that effective query optimization increasingly requires cross-modal reasoning as user queries become more diverse.

\textbf{Remark on Cross-Category Methods.} Several methods in this subsection, including CRAG, Adaptive-RAG, Speculative RAG, and Think-then-Act, are general-purpose RAG improvement methods rather than pure disambiguation techniques. We discuss them here because their core mechanisms involve \textit{implicit disambiguation}: they assess and route queries based on confidence signals, effectively disambiguating between cases where different processing strategies are appropriate. This reflects the broader trend toward adaptive systems that blur the boundaries between distinct optimization operations.

\textsc{CRAG}~\citep{CRAG} (Corrective Retrieval Augmented Generation) introduces a lightweight retrieval evaluator that assesses the overall quality of retrieved documents and returns a confidence score. Based on this score, the system triggers different knowledge retrieval actions: when retrieval quality is high, the documents are used directly; when quality is low, CRAG activates web search as an expanded knowledge source. A decompose-then-recompose algorithm selectively focuses on key information within retrieved documents while filtering out irrelevant content. CRAG is plug-and-play and can be seamlessly integrated with various RAG methods, representing an important feedback-driven approach to dynamically correcting retrieval quality.

\textsc{Adaptive-RAG}~\citep{AdaptiveRAG} proposes learning to adapt retrieval strategies based on question complexity. The system uses a classifier trained to categorize queries into different complexity levels (simple, moderate, complex) and routes each query to the most appropriate processing pipeline: no retrieval for simple factoid questions, single-step retrieval for moderate queries, or multi-step iterative retrieval for complex ones. This adaptive routing approach ensures that computational resources are allocated proportionally to query difficulty, avoiding unnecessary overhead for simple queries while providing sufficient depth for complex ones.

\textsc{Speculative RAG}~\citep{SpeculativeRAG} enhances retrieval-augmented generation through a drafting mechanism that leverages a smaller, distilled specialist LLM to generate multiple draft responses in parallel from different subsets of retrieved documents. A larger, generalist LLM then efficiently verifies these drafts, selecting the best candidate. This approach reduces latency by parallelizing the generation of candidate answers while maintaining quality through verification, providing an effective feedback-driven disambiguation strategy when multiple retrieval results suggest different interpretations.

\textsc{Think-then-Act}~\citep{Think-then-Act} proposes a dual-angle evaluation framework for retrieval-augmented generation that assesses both the relevance of retrieved documents and the quality of generated responses. By introducing feedback from both retrieval evaluation and generation evaluation, the system learns to adaptively adjust its query processing strategy, disambiguating between cases where retrieval failure versus generation failure is the primary bottleneck.

\textsc{RARG}~\citep{RARG} (Retrieval Augmented Response Generation) applies evidence-driven retrieval to address misinformation, where the system must disambiguate between accurate and misleading information in retrieved results. The evidence-based verification mechanism provides a principled approach to query disambiguation in adversarial retrieval settings.

\begin{table}[t]
    \centering
    \scriptsize
    \caption{Comparative analysis of query disambiguation methods.}
    \label{tab:disambiguation}
    \adjustbox{max width=\columnwidth}{%
    \begin{tabular}{llccc}
    \toprule
    \textbf{Method} & \textbf{Approach} & \textbf{Multi-Turn} & \textbf{Feedback} & \textbf{Year} \\
    \midrule
    ToC & Tree clarification & No & External KB & 2023 \\
    EchoPrompt & Rephrasing & No & None & 2024 \\
    InfoCQR & Rewrite-then-edit & No & None & 2023 \\
    Rewrite-Ret.-Read & RL-trained rewriting & No & Retrieval & 2023 \\
    AdaQR & Preference optimization & Yes & Retrieval & 2024 \\
    MaFeRw & Multi-aspect feedback & No & Retrieval+Gen & 2024 \\
    RQ-RAG & Learned rewriting & No & Retrieval & 2024 \\
    DMQR-RAG & Multi-query rewriting & No & Retrieval & 2024 \\
    RankRAG & Unified rank+gen & No & Ranking & 2024 \\
    CHIQ & Conversational & Yes & None & 2024 \\
    LLM4CS & Prompting framework & Yes & Retrieval & 2023 \\
    GuideCQR & Doc-guided rewriting & Yes & Retrieval & 2024 \\
    RaFe & Ranking feedback & No & Ranking & 2024 \\
    ERRR & Evaluate-then-rewrite & No & Retrieval & 2024 \\
    OmniSearch & Multi-modal disambig. & Yes & Vision+Text & 2024 \\
    CRAG & Corrective retrieval & No & Retrieval eval. & 2024 \\
    Adaptive-RAG & Complexity routing & No & Classifier & 2024 \\
    Speculative RAG & Draft verification & No & LLM verify & 2024 \\
    Think-then-Act & Dual-angle eval. & No & Dual feedback & 2024 \\
    RARG & Evidence-driven & No & Verification & 2024 \\
    \bottomrule
    \end{tabular}%
    }
\end{table}

\textbf{Evolution and Synthesis.} Disambiguation methods span a spectrum from explicit clarification (ToC, EchoPrompt) to implicit, feedback-driven approaches (AdaQR, MaFeRw, RAG-DDR). A notable trend is the convergence of disambiguation with \textit{adaptive routing}: methods like CRAG, Adaptive-RAG, and Think-then-Act do not disambiguate the query text itself, but rather disambiguate the \textit{processing strategy}, determining which pipeline component is the bottleneck and routing accordingly. This suggests that future disambiguation systems will increasingly operate at the meta-level, choosing among optimization strategies rather than directly modifying queries. The conversational disambiguation subcategory (CHIQ, LLM4CS, GuideCQR) highlights an important underexplored direction: leveraging dialogue history as a rich disambiguation signal, which may reduce the need for explicit clarification questions.

\section{Query Abstraction}\label{sec:abstraction}

For complex multi-hop queries, sequential decomposition may introduce errors that compound through the reasoning chain. Query abstraction addresses this challenge by elevating queries to higher-level conceptual representations that capture underlying principles or patterns. This approach reduces the chance of making errors in intermediate reasoning steps by operating at a level where the reasoning structure is clearer and more robust. We organize abstraction methods into conceptual abstraction (which identifies higher-level principles) and pattern-based abstraction (which extracts generalizable reasoning structures). Table~\ref{tab:abstraction} provides a comparative overview of representative methods.

\vspace{0.3em}
\noindent\fbox{\parbox{0.96\columnwidth}{
\textbf{Key Insight: The Abstraction-Grounding Duality.} Abstraction methods like Step-Back demonstrate a paradox: moving \textit{away} from the specific query (to general principles) improves \textit{specific} answer quality. This suggests that LLMs possess implicit knowledge organized hierarchically: accessing high-level concepts activates relevant lower-level details. The practical implication: for complex queries, ``zoom out before zooming in'' is often more effective than direct retrieval.
}}
\vspace{0.3em}

\subsection{Conceptual Abstraction: Step-Back Reasoning}

Conceptual abstraction methods prompt models to consider the broader context or principles behind specific queries before addressing the specifics.

\subsubsection{Foundational Step-Back Methods}

\textsc{Step-Back}~\citep{StepBack} prompting is the foundational method for conceptual abstraction in query optimization. The approach prompts LLMs to ``step back'' from the specific query and consider the broader principles or concepts involved. By first identifying the high-level category or framework relevant to the query, the model can then apply this abstracted knowledge to address the specific instance. For example, when asked about a specific chemical reaction, the model first retrieves general principles of chemistry that govern such reactions, then applies these principles to the specific case.

This approach is particularly effective for queries requiring domain expertise or principled reasoning, where direct retrieval of specific facts may be insufficient. The abstracted principles provide a reasoning scaffold that guides the model toward correct answers even when specific facts are not directly retrievable.

\subsubsection{Abstract Variable and Skeleton Methods}

\textsc{Chain-of-Abstraction (CoA)}~\citep{CoA} abstracts general chain-of-thought reasoning into chains with abstract variables. This enables LLMs to solve queries by utilizing domain-specialized tools, such as calculators for mathematical operations or search engines for factual lookups. The abstract variables serve as placeholders for tool outputs, allowing the reasoning chain to be constructed before specific values are computed. This separation of reasoning structure from specific values provides robustness against computational errors and enables verification of reasoning logic independent of specific facts.

\textsc{Abstraction-of-Thought (AoT)}~\citep{AoT} uses an abstract skeletal framework to structure the entire reasoning process. Unlike unconstrained chain-of-thought, AoT explicitly integrates different levels of abstraction throughout reasoning. At each higher level, the abstraction is a distilled version containing fewer concrete details while clearly stating the objective and functionality of each reasoning step. This hierarchical structure makes reasoning more transparent and enables identification of errors at appropriate abstraction levels.

\subsubsection{Abstraction Training and Evaluation}

\textsc{AbsInstruct}~\citep{AbsInstruct} elicits abstraction ability through explanation tuning with plausibility estimation, teaching models to generate and evaluate abstract representations. The training approach explicitly develops the model's capacity for abstraction, rather than relying solely on prompting.

\textsc{AbsPyramid}~\citep{AbsPyramid} provides a benchmark for evaluating abstraction ability using a unified entailment graph. The benchmark enables systematic comparison of different abstraction methods and identification of capability gaps.

\subsection{Pattern-Based Abstraction: Rule and Structure Learning}

Pattern-based abstraction methods extract generalizable reasoning patterns or structural templates that can be applied across different queries.

\subsubsection{Meta-Reasoning Approaches}

\textsc{Meta-Reasoning}~\citep{Meta-Reasoning} deconstructs the semantics of entities and operations into generic symbolic representations. This methodology allows LLMs to learn generalized reasoning patterns across semantically diverse scenarios. By abstracting from specific entities to generic roles, the model can apply learned patterns to novel situations that share structural similarity with training examples.

\textsc{Conceptualization-Abstraction}~\citep{Conceptualization-Abstraction} investigates how language models can perform conceptual and unbiased reasoning by mapping specific instances to abstract conceptual categories. The work demonstrates that explicit conceptualization, i.e., replacing specific entities with their conceptual types, enables more robust and generalizable reasoning, particularly in scenarios involving knowledge-intensive inference where surface-level features may mislead.

\subsubsection{Rule-Guided Methods}

\textsc{RuleRAG}~\citep{RuleRAG} observes that widespread logical rules can guide task completion. The method recalls documents supporting queries logically in the directions of explicit rules, generating final responses based on both retrieved information and attributable rules. This rule-guided approach provides interpretable reasoning that can be verified against known principles.

\textsc{Crafting-the-Path}~\citep{Crafting-the-Path} introduces a structured query rewriting approach that extracts higher-level conceptual understanding through query concept identification, query type recognition, and expected answer extraction. By reducing dependence on LLM parametric knowledge, this method generates queries that are both more robust and contain fewer factual errors, particularly in domains unfamiliar to the model.

\subsubsection{Graph-Based Abstraction}

Graph-based methods occupy a unique position in our taxonomy: while they are often characterized as retrieval infrastructure, their core contribution to query optimization lies in \textit{structural abstraction}: transforming flat textual queries into graph-structured representations that capture entities, relationships, and hierarchical concepts. We categorize them under abstraction because the query transformation (text $\rightarrow$ graph pattern) fundamentally elevates the query to a more structured, abstract representation.

\textsc{SimGRAG}~\citep{SimGRAG} addresses query-knowledge graph alignment through a two-stage process. First, queries are transformed into desired graph patterns using LLMs. Second, alignment between patterns and candidate subgraphs is quantified using graph semantic distance (GSD). This approach abstracts queries into structural patterns that can be matched against knowledge graph structures.

\textsc{LPKG}~\citep{LPKG} (Learning to Plan from Knowledge Graphs) takes a complementary approach by learning retrieval planning strategies directly from knowledge graph structures. The method trains models to generate structured retrieval plans that specify which entities and relations to traverse, effectively abstracting complex queries into graph navigation strategies. This planning-based abstraction enables systematic exploration of knowledge graphs for multi-hop reasoning.

\textsc{MA-RIR}~\citep{MA-RIR} (Multi-Aspect Retrieval for Information Reasoning) defines query aspects as sub-spans representing distinct topics or facets. This facet-based abstraction enables focused reasoning across different aspects of complex queries.

\textsc{GraphRAG}~\citep{GraphRAG} represents a significant advancement in combining knowledge graphs with retrieval-augmented generation. Unlike traditional RAG that retrieves flat document chunks, GraphRAG constructs a hierarchical knowledge graph from source documents, extracting entities and relationships to form a structured representation. Query processing involves matching the query against this graph structure, enabling multi-hop reasoning across connected entities. GraphRAG excels at global summarization queries and questions requiring synthesis across multiple documents, addressing a key limitation of chunk-based retrieval.

\textsc{LightRAG}~\citep{LightRAG} introduces a lightweight and efficient graph-based retrieval framework that addresses the high computational overhead of full knowledge graph construction in GraphRAG. LightRAG employs an incremental graph update algorithm and a dual-level retrieval paradigm that operates at both local (entity-specific) and global (topic-level) granularities. By combining low-level entity extraction with high-level keyword-driven graph traversal, LightRAG achieves competitive or superior performance to GraphRAG while significantly reducing indexing and query-time costs. This demonstrates that abstraction through lightweight graph structures can balance quality and efficiency for large-scale knowledge bases.

\textsc{MemoRAG}~\citep{MemoRAG} proposes a memory-inspired knowledge discovery paradigm that employs a dual-system architecture for query abstraction. A lightweight, long-context LLM forms a global memory over the entire database, serving as a compressed abstraction of the knowledge space. When a query is posed, this memory model generates draft answers and retrieval clues that capture the high-level intent and relevant knowledge areas, effectively abstracting the query into a richer representation. These clues then guide a fine-grained retriever to locate specific evidence from the database, and a more powerful LLM generates the final answer. MemoRAG extends standard RAG to handle queries with implicit or ambiguous information needs where direct retrieval may fail, demonstrating that global memory serves as an effective form of knowledge abstraction.

\subsubsection{Heterogeneous Document Abstraction}

\textsc{TableRAG}~\citep{TableRAG} addresses the challenging problem of retrieval-augmented generation over heterogeneous documents containing both textual and tabular components. The method introduces specialized abstraction mechanisms for table understanding, including cell-level semantic parsing and cross-modal alignment between table headers and query terms. TableRAG demonstrates that treating tables as first-class citizens in the retrieval process, rather than flattening them to text, significantly improves performance on document QA tasks involving structured data.

\begin{table}[t]
    \centering
    \scriptsize
    \caption{Comparative analysis of query abstraction methods.}
    \label{tab:abstraction}
    \adjustbox{max width=\columnwidth}{%
    \begin{tabular}{llccc}
    \toprule
    \textbf{Method} & \textbf{Abstraction Type} & \textbf{Tool Use} & \textbf{Structure} & \textbf{Year} \\
    \midrule
    Step-Back & Conceptual & No & Principles & 2024 \\
    CoA & Variable abstraction & Yes & Chain & 2024 \\
    AoT & Hierarchical skeleton & No & Tree & 2024 \\
    AbsInstruct & Learned abstraction & No & N/A & 2024 \\
    Meta-Reasoning & Symbolic patterns & No & Generic & 2024 \\
    Concept.-Abstract. & Conceptual mapping & No & Categorical & 2024 \\
    RuleRAG & Rule-guided & No & Logical & 2024 \\
    SimGRAG & Graph patterns & No & Graph & 2025 \\
    LPKG & KG-based planning & Yes & Graph & 2024 \\
    GraphRAG & Knowledge graph & Yes & Graph & 2024 \\
    LightRAG & Lightweight graph & No & Graph & 2025 \\
    MemoRAG & Memory-inspired & No & Global memory & 2024 \\
    TableRAG & Heterogeneous doc & Yes & Table+Text & 2025 \\
    \bottomrule
    \end{tabular}%
    }
\end{table}

\textbf{Evolution and Synthesis.} Abstraction methods reveal an important design principle: \textit{the level of abstraction should match query complexity}. For moderately complex queries, lightweight abstraction (Step-Back prompting) suffices; for queries requiring cross-document synthesis, graph-based abstractions (GraphRAG, LightRAG) provide the necessary structural scaffolding; and for queries with entirely implicit information needs, memory-based abstractions (MemoRAG) offer global knowledge compression. The graph-based methods in particular represent a convergence point between abstraction and structured retrieval, suggesting that the boundary between ``query transformation'' and ``retrieval architecture'' is increasingly porous. An open challenge remains: current abstraction methods largely operate at a single level, whereas human reasoning naturally shifts between multiple abstraction levels, a capability that future methods should aspire to replicate.

\section{Evaluation and Benchmarks}\label{sec:evaluation}

The evaluation of query optimization techniques presents unique challenges due to the indirect relationship between query quality and final response correctness. A well-optimized query that retrieves relevant documents may still lead to incorrect answers due to LLM limitations, while a poorly optimized query may occasionally succeed through luck. This section examines evaluation methodologies, benchmarks, and the challenges facing systematic assessment.

\subsection{Evaluation Dimensions}

Query optimization can be evaluated along multiple complementary dimensions:

\textbf{Retrieval Quality Metrics.} These metrics assess whether optimized queries retrieve more relevant documents:
\begin{itemize}[leftmargin=*,nosep]
    \item \textit{Recall@K}: Fraction of relevant documents appearing in top-K results
    \item \textit{Mean Reciprocal Rank (MRR)}: Average inverse rank of first relevant document
    \item \textit{Normalized Discounted Cumulative Gain (nDCG)}: Graded relevance measure accounting for rank position
    \item \textit{Precision@K}: Fraction of top-K results that are relevant
\end{itemize}

\textbf{End-to-End Performance Metrics.} These metrics assess whether query optimization improves final answer quality:
\begin{itemize}[leftmargin=*,nosep]
    \item \textit{Exact Match (EM)}: Binary correctness of extracted answers
    \item \textit{F1 Score}: Token-level overlap between predicted and reference answers
    \item \textit{Accuracy}: Correctness rate on downstream QA benchmarks
    \item \textit{ROUGE/BLEU}: N-gram overlap for generation tasks
\end{itemize}

\textbf{Efficiency Metrics.} These metrics assess practical deployment costs:
\begin{itemize}[leftmargin=*,nosep]
    \item \textit{LLM API Calls}: Number of calls to language model APIs
    \item \textit{Retrieval Latency}: Time for retrieval operations
    \item \textit{Total Wall-Clock Time}: End-to-end processing time
    \item \textit{Token Usage}: Input/output tokens consumed
\end{itemize}

\textbf{Robustness Metrics.} These metrics assess generalization:
\begin{itemize}[leftmargin=*,nosep]
    \item \textit{Cross-Domain Transfer}: Performance across different domains
    \item \textit{Retriever Agnosticism}: Consistency across different retrieval systems
    \item \textit{Query Complexity Scaling}: Performance across complexity classes
\end{itemize}

\subsection{Benchmark Landscape}

Several benchmark categories have been used to evaluate query optimization:

\textbf{Single-Hop QA Benchmarks.} These assess performance on straightforward factoid queries:
\begin{itemize}[leftmargin=*,nosep]
    \item Natural Questions~\citep{DBLP:journals/tacl/KwiatkowskiPRCP19}: Real Google search queries
    \item TriviaQA~\citep{DBLP:conf/acl/JoshiCWZ17}: Trivia enthusiast questions
    \item WebQuestions~\citep{WebQuestions}: Questions answerable from Freebase
\end{itemize}

\textbf{Multi-Hop QA Benchmarks.} These evaluate complex reasoning requiring multiple evidence sources:
\begin{itemize}[leftmargin=*,nosep]
    \item HotpotQA~\citep{DBLP:conf/emnlp/Yang0ZBCSM18}: Multi-hop questions with supporting facts
    \item 2WikiMultiHopQA~\citep{DBLP:conf/coling/HoNSA20}: Cross-document reasoning
    \item MuSiQue~\citep{DBLP:journals/tacl/TrivediBKS22}: Multi-step reasoning with compositional questions
\end{itemize}

\textbf{Conversational QA Benchmarks.} These test disambiguation in multi-turn dialogue:
\begin{itemize}[leftmargin=*,nosep]
    \item QReCC~\citep{DBLP:conf/naacl/AnanthaRTSCC21}: Conversational question reformulation
    \item TopiOCQA~\citep{DBLP:journals/tacl/AdlakhaSKCR22}: Topic-oriented conversational QA
\end{itemize}

\textbf{RAG-Specific Benchmarks.} These provide comprehensive evaluation of RAG system components:
\begin{itemize}[leftmargin=*,nosep]
    \item RAD-Bench~\citep{rag_benchmark_1}: RAG dialogue benchmark with retrieval quality evaluation
    \item RAG-QA Arena~\citep{rag_benchmark_3}: Domain-robust long-form QA evaluation
    \item Sub-question coverage benchmarks~\citep{rag_benchmark_2}: Evaluating response completeness via sub-question decomposition
\end{itemize}

\subsection{Evaluation Challenges and Gaps}

Despite progress, significant evaluation challenges remain:

\textbf{Lack of Query-Level Annotations.} Most benchmarks provide only end-to-end labels without ground truth for intermediate query transformations. This makes it difficult to assess whether optimization methods are improving queries for the right reasons or succeeding through confounding factors.

\textbf{Retriever Dependence.} Query optimization effectiveness varies significantly across different retrieval systems (sparse vs.\ dense, different embedding models). Methods optimized for one retriever may not transfer well to others, complicating fair comparison.

\textbf{Efficiency-Quality Trade-offs.} Current benchmarks rarely provide standardized efficiency measurements, making it difficult to assess practical deployment costs and compare methods on Pareto frontiers of quality and efficiency.

\textbf{Limited Coverage of Complex Scenarios.} Most benchmarks focus on single-turn factoid questions, with limited coverage of multi-modal queries, conversational contexts, and domain-specific applications.

\textbf{Cross-Paper Comparison Difficulties.} Due to varying experimental setups, model sizes, retrieval corpora, and evaluation protocols across different papers, direct numerical comparisons across methods are inherently unreliable. Key confounding factors include: (1) \textit{base LLM variation}: methods evaluated with GPT-4 vs.\ LLaMA-7B yield incomparable numbers; (2) \textit{retrieval corpus differences}: Wikipedia snapshots, web corpora, and domain-specific collections produce different retrieval quality; (3) \textit{evaluation metric inconsistencies}: some papers report Exact Match (EM), others report F1 or accuracy, often without specifying tokenization details; and (4) \textit{prompt and few-shot configuration}: the number and choice of demonstrations significantly affect results. We therefore deliberately refrain from presenting a unified performance comparison table, as doing so would create a misleading impression of commensurability. Instead, we encourage readers to consult the original papers and, where possible, reproduce results under controlled settings. Standardized evaluation frameworks such as RAGAS~\citep{RAGAS} and ARES~\citep{ARES} represent promising steps toward enabling fairer cross-method comparisons.

\section{Discussion and Practical Guidance}\label{sec:discussion}

Having surveyed the landscape of query optimization techniques, we now provide comparative analysis and practical guidance to help researchers and practitioners navigate the complex design space.

\subsection{Emergent Design Principles}

From our comprehensive analysis, we distill five emergent design principles that characterize successful query optimization systems:

\vspace{0.3em}
\noindent\fbox{\parbox{0.96\columnwidth}{
\textbf{Design Principle 1: Adaptive Complexity Allocation.} 
The most effective systems dynamically allocate computational effort based on query difficulty. Simple factoid queries warrant minimal processing (single-pass expansion), while complex analytical queries justify substantial investment (multi-step decomposition + abstraction). Self-RAG, DeepRAG, and RAG-Gym exemplify this principle through learned decision policies. Notably, both DeepRAG and RAG-Gym formalize retrieval decisions as MDPs, suggesting that formal optimization frameworks provide principled approaches to adaptive allocation.

\textbf{Design Principle 2: Feedback-Driven Refinement.}
Top-performing methods create closed loops between query optimization and downstream outcomes. RAG-DDR's differentiable rewards and MaFeRw's multi-aspect feedback demonstrate that end-to-end learning signals significantly outperform heuristic optimization.

\textbf{Design Principle 3: Compositional Operation Design.}
Query optimization operations should be designed for composition rather than isolation. Methods combining expansion + decomposition or disambiguation + abstraction consistently outperform single-operation approaches on complex benchmarks.

\textbf{Design Principle 4: Explicit vs.\ Implicit Reasoning Trade-offs.}
Explicit reasoning traces (ReAct, CoT) improve interpretability and debugging but increase latency. Implicit reasoning (HyDE, learned embeddings) offers efficiency but sacrifices transparency. Production systems should balance based on domain requirements.

\textbf{Design Principle 5: Retriever-Aware Optimization.}
Query transformations should account for the target retriever's characteristics. Dense retrievers benefit more from semantic expansion (HyDE), while sparse retrievers respond better to keyword augmentation (Query2Doc). Retriever-agnostic methods often underperform specialized approaches.
}}
\vspace{0.3em}

\subsection{Synthesis: Patterns Across the Field}

Beyond individual design principles, we identify four overarching patterns that emerge across the methods reviewed:

\vspace{0.3em}
\noindent\fbox{\parbox{0.96\columnwidth}{
\textbf{Pattern 1: From Retrieval-Centric to Reasoning-Centric.}
Early methods (2020-2022) treated query optimization as a retrieval preprocessing step. Contemporary methods increasingly view it as a \textit{reasoning task} where the LLM actively decides what information it needs. This shift reflects the growing capabilities of frontier models.

\textbf{Pattern 2: The Rise of Process Supervision.}
Methods have evolved from outcome-only evaluation (did we get the right answer?) to process supervision (did each step make sense?). RAG-Gym's MDP formulation, RAG-Star's MCTS, DeepRAG's chain-of-calibration, and CoRAG's learnable retrieval chains represent this trend toward fine-grained optimization signals that reward intermediate reasoning quality.

\textbf{Pattern 3: Convergence Toward Agentic Architectures.}
Distinct optimization operations (expansion, decomposition, disambiguation, abstraction) are converging into unified agentic frameworks where models autonomously select and apply appropriate strategies. Agentic-RAG and OmniSearch exemplify this integration, while DeepRAG demonstrates that even expansion-focused methods increasingly incorporate decomposition elements, and MemoRAG shows that abstraction can be combined with implicit disambiguation through global memory.

\textbf{Pattern 4: Multi-Modal Expansion.}
Query optimization is rapidly extending beyond text. TableRAG, GraphRAG, and OmniSearch indicate that future systems must handle heterogeneous query types spanning text, tables, images, and structured data.
}}
\vspace{0.3em}

\subsection{Comparative Analysis of Operations}

\begin{table*}[t]
    \centering
    \scriptsize
    \caption{Comprehensive comparison of query optimization operations across multiple dimensions.}
    \label{tab:comprehensive}
    \resizebox{\textwidth}{!}{%
    \begin{tabular}{p{1.6cm}p{2.6cm}p{2.6cm}p{2.6cm}p{2.6cm}}
    \toprule
    \textbf{Dimension} & \textbf{Expansion} & \textbf{Decomposition} & \textbf{Disambiguation} & \textbf{Abstraction} \\
    \midrule
    \textbf{Primary Goal} & Improve recall via semantic enrichment & Divide-and-conquer for multi-faceted queries & Resolve ambiguity, clarify intent & Principled reasoning via abstractions \\
    \midrule
    \textbf{Best For} & Simple factoid queries & Multi-hop reasoning queries & Ambiguous queries & Complex analytical queries \\
    \midrule
    \textbf{Latency} & Low-Medium & High (multiple retrievals) & Medium & Medium \\
    \midrule
    \textbf{Error Risk} & Low & High (cascading) & Medium & Medium \\
    \midrule
    \textbf{LLM Dep.} & Medium & High & High & High \\
    \midrule
    \textbf{Limitation} & May reduce precision & Error cascade & Detection is hard & May oversimplify \\
    \bottomrule
    \end{tabular}%
    }
\end{table*}

Table~\ref{tab:comprehensive} provides a multi-dimensional comparison of the four fundamental operations. Key insights include:

\textbf{Trade-offs Between Recall and Precision.} Expansion techniques primarily improve recall at the potential cost of precision. Disambiguation, conversely, focuses on improving precision by narrowing down intent. Practitioners must balance these trade-offs based on application requirements.

\textbf{Latency Considerations.} Decomposition introduces the highest latency due to multiple retrieval operations, making it less suitable for real-time applications. Expansion with single-pass retrieval offers the best latency characteristics.

\textbf{Error Propagation Patterns.} Sequential decomposition is particularly vulnerable to error cascades, where mistakes in early sub-queries compound through the reasoning chain. Parallel decomposition and expansion provide better error isolation.

\subsection{Case Studies and Examples}

To illustrate the practical application of different optimization strategies, we present representative case studies across query complexity classes in Table~\ref{tab:cases}.

\begin{table}[t]
    \centering
    \scriptsize
    \caption{Illustrative case studies showing how different query types benefit from appropriate optimization strategies.}
    \label{tab:cases}
    \resizebox{\columnwidth}{!}{
    \begin{tabular}{p{1.5cm}p{4cm}p{3.5cm}p{4cm}}
    \toprule
    \textbf{Class} & \textbf{Example Query} & \textbf{Recommended Strategy} & \textbf{Optimization Result} \\
    \midrule
    Class I & ``Who founded Tesla?'' & Expansion (HyDE) & Generate hypothetical: ``Tesla was founded by Elon Musk...'' $\rightarrow$ improved embedding similarity \\
    \midrule
    Class II & ``Compare GPT-4 and Claude on coding tasks'' & Decomposition (Plan×RAG) & Sub-queries: (1) GPT-4 coding performance; (2) Claude coding performance; (3) Comparative analysis \\
    \midrule
    Class III & ``Is this investment safe?'' [with context] & Disambiguation (ToC) & Clarify: Financial safety? Legal safety? Generate tree of interpretations \\
    \midrule
    Class IV & ``How will quantum computing affect cybersecurity?'' & Abstraction (Step-Back) & First retrieve: ``Principles of quantum cryptography'' $\rightarrow$ then apply to specific question \\
    \bottomrule
    \end{tabular}
    }
\end{table}

\subsubsection{Case Study 1: Simple Factoid Query (Class I)}

\textbf{Query}: ``What programming language was used to build Instagram's backend?''

\textbf{Challenge}: The query is straightforward but may suffer from vocabulary mismatch; relevant documents might not contain the exact phrase ``programming language'' but instead mention ``tech stack'' or ``built with.''

\textbf{Solution}: Apply HyDE to generate a hypothetical document: ``Instagram's backend was built using Django, a Python web framework. The company chose Python for its development speed and Django's robust features for handling high traffic...''

\textbf{Outcome}: The hypothetical document provides semantic bridges (Django, Python, web framework) that improve retrieval of relevant technical articles.

\textbf{Analysis}: This case illustrates the \textit{Semantic Signature Principle} discussed in Section~\ref{sec:expansion}: even if the generated document contains inaccuracies (e.g., wrong version numbers), the embedding captures the right topical neighborhood. However, applying decomposition or abstraction here would be counterproductive, as the query is atomic and the evidence is explicit, validating our Class~I $\rightarrow$ Expansion mapping.

\subsubsection{Case Study 2: Multi-Hop Reasoning Query (Class II)}

\textbf{Query}: ``Which country had higher GDP growth in 2023: the country where the World Wide Web was invented or the largest producer of rare earth elements?''

\textbf{Challenge}: This query requires resolving two implicit references (``country where the WWW was invented'' $\rightarrow$ Switzerland/CERN or UK, ``largest rare earth producer'' $\rightarrow$ China) and then comparing economic data.

\textbf{Solution}: Apply sequential decomposition with Self-Ask:
\begin{enumerate}[leftmargin=*,nosep]
    \item ``Where was the World Wide Web invented?'' $\rightarrow$ CERN (Switzerland), by Tim Berners-Lee (British)
    \item ``Which country produces the most rare earth elements?'' $\rightarrow$ China
    \item Retrieve and compare GDP growth data for these countries
\end{enumerate}

\textbf{Outcome}: Sequential decomposition correctly resolves implicit references and retrieves necessary factual data for comparison. Note that the first sub-query itself reveals ambiguity (Switzerland vs.\ UK), demonstrating how decomposition and disambiguation can interact.

\textbf{Analysis}: This case highlights the \textit{Error Propagation-Parallelism Trade-off} from Section~\ref{sec:decomposition}: the sub-queries have genuine sequential dependencies (the third step depends on results of the first two), mandating sequential execution. A parallel approach (Plan×RAG) could execute the first two sub-queries concurrently, substantially reducing latency for the initial resolution phase, while reserving sequential execution only for the final comparison step.

\subsubsection{Case Study 3: Ambiguous Query (Class III)}

\textbf{Query}: ``Is Apple a good investment?''

\textbf{Challenge}: Multiple ambiguities exist: (1) ``Apple'' could refer to the company or the fruit commodity market; (2) ``good'' is subjective and depends on investment horizon, risk tolerance, and current portfolio.

\textbf{Solution}: Apply Tree of Clarifications (ToC):
\begin{itemize}[leftmargin=*,nosep]
    \item Branch 1: Apple Inc. stock $\rightarrow$ Short-term trading vs.\ Long-term holding
    \item Branch 2: Apple Inc. stock $\rightarrow$ Growth investors vs.\ Value investors
    \item Branch 3: Apple commodity $\rightarrow$ Agricultural investment perspective
\end{itemize}

\textbf{Outcome}: The system generates a comprehensive response addressing multiple interpretations, or prompts the user for clarification if interaction is possible.

\textbf{Analysis}: This case exemplifies the Key Insight from Section~\ref{sec:disambiguation}: ambiguity is often a feature, not a bug. Rather than aggressively disambiguating to a single interpretation, ToC's multi-branch exploration provides more complete coverage. In a production system, the choice between exploration (generating multi-branch answers) and clarification (asking the user) depends on interaction affordances: conversational systems can ask, while single-turn systems should explore.

\subsubsection{Case Study 4: Complex Analytical Query (Class IV)}

\textbf{Query}: ``How might the adoption of autonomous vehicles affect urban planning in the next 20 years?''

\textbf{Challenge}: This query requires synthesizing knowledge across multiple domains (autonomous vehicle technology, urban planning, transportation economics, social behavior) with long-horizon predictive reasoning.

\textbf{Solution}: Apply Step-Back abstraction:
\begin{enumerate}[leftmargin=*,nosep]
    \item Identify higher-level principles: ``What are the general principles by which transportation technology changes affect urban development?''
    \item Retrieve relevant historical patterns: Automobile adoption impact, rail transit effects
    \item Apply principles to autonomous vehicles: Reduced parking needs, changed commute patterns, suburban expansion potential
\end{enumerate}

\textbf{Outcome}: The abstraction step provides a principled framework for reasoning about a novel, speculative question by grounding it in established urban planning theory.

\textbf{Analysis}: This case demonstrates the \textit{Abstraction-Grounding Duality} from Section~\ref{sec:abstraction}: by first retrieving general principles (how past transportation revolutions shaped cities), the system obtains a reasoning scaffold that makes the specific, speculative question tractable. Direct decomposition would struggle here because the sub-questions are not well-defined a priori; it is the abstract principles that reveal \textit{which} sub-questions to ask (parking infrastructure, commute patterns, land use).

\subsection{Decision Framework for Strategy Selection}

Based on our analysis, we propose a practical decision framework for selecting query optimization strategies:

\begin{figure}[t]
\centering
\resizebox{0.58\columnwidth}{!}{%
\begin{tikzpicture}[
    decision/.style={diamond, draw, fill=yellow!20, text width=1.2cm, align=center, inner sep=1pt, aspect=2, font=\scriptsize},
    process/.style={rectangle, draw, fill=blue!12, text width=1.1cm, align=center, rounded corners=2pt, font=\scriptsize, minimum height=0.45cm, line width=0.4pt},
    outcome/.style={rectangle, draw, rounded corners=2pt, text width=1.1cm, align=center, font=\scriptsize, minimum height=0.45cm, line width=0.4pt},
    arrow/.style={->,>=stealth, line width=0.5pt}]
    
    \node[process] (start) at (0,0) {Analyze\\Query};
    
    \node[decision] (d1) at (0,-1.1) {Ambig.?};
    
    \node[outcome, fill=disambig-color!40] (disamb) at (1.8,-1.1) {Disambig.};
    
    \node[decision] (d2) at (-1.4,-2.4) {Multi-hop?};
    
    \node[decision] (d3) at (1.8,-2.4) {Concept.?};
    
    \node[outcome, fill=decomp-color!40] (decomp) at (-2.4,-3.6) {Decomp.};
    \node[outcome, fill=expansion-color!40] (expand1) at (-0.5,-3.6) {Expansion};
    \node[outcome, fill=expansion-color!40] (expand2) at (1.1,-3.6) {Expansion};
    \node[outcome, fill=abstract-color!40] (abstr) at (2.7,-3.6) {Abstract.};
    
    \draw[arrow] (start) -- (d1);
    \draw[arrow] (d1) -- node[above, font=\tiny] {Y} (disamb);
    \draw[arrow] (d1) -- node[left, font=\tiny] {N} (d2);
    \draw[arrow] (disamb) -- (d3);
    \draw[arrow] (d2) -- node[left, font=\tiny] {Y} (decomp);
    \draw[arrow] (d2) -- node[right, font=\tiny] {N} (expand1);
    \draw[arrow] (d3) -- node[left, font=\tiny] {N} (expand2);
    \draw[arrow] (d3) -- node[right, font=\tiny] {Y} (abstr);
\end{tikzpicture}%
}
\caption{Decision flowchart for query optimization strategy selection.}
\label{fig:decision}
\end{figure}

Figure~\ref{fig:decision} presents a simplified decision flowchart. The key decision points are:

\begin{enumerate}[leftmargin=*,nosep]
    \item \textbf{Ambiguity Check}: If the query contains ambiguous terms, multiple interpretations, or underspecified context, prioritize disambiguation before other operations.
    \item \textbf{Complexity Assessment}: For unambiguous queries, assess whether the query requires information from multiple sources (multi-hop) or can be answered from a single source.
    \item \textbf{Reasoning Type}: For queries requiring synthesis, determine whether the reasoning is primarily factual aggregation (decomposition) or conceptual/analytical (abstraction).
\end{enumerate}

\subsection{Implementation Recommendations}

Based on our survey, we offer the following implementation recommendations for practitioners:

\textbf{Start Simple.} For many applications, simple expansion techniques (e.g., HyDE, Query2Doc) provide substantial improvements with minimal complexity. Only introduce more sophisticated methods when simpler approaches prove insufficient.

\textbf{Profile Query Distribution.} Analyze the distribution of query types in your application. If most queries are simple factoid questions (Class I), invest in robust expansion. If multi-hop queries dominate (Class II), prioritize decomposition infrastructure. Recent work on unsupervised query routing~\citep{UnsupervisedQueryRouting} demonstrates that clustering-based approaches can automatically learn to route queries to the most appropriate optimization pipeline without requiring labeled training data. By grouping similar queries and learning routing policies from retrieval feedback, such methods provide a practical mechanism for adaptive strategy selection at scale.

\textbf{Consider Latency Constraints.} For real-time applications, prefer single-pass methods (expansion) over iterative methods (sequential decomposition). Batch or offline applications can tolerate higher latency for better quality.

\textbf{Build Modular Pipelines.} Design systems that can compose multiple operations. A query might benefit from expansion \textit{after} decomposition, or disambiguation \textit{before} abstraction.

\textbf{Monitor and Adapt.} Query optimization effectiveness varies across domains and user populations. Implement monitoring to track optimization quality and adapt strategies based on observed performance.

\section{Challenges and Future Directions}\label{sec:challenges}

Based on our comprehensive analysis of the query optimization landscape, building on the patterns identified in Section~\ref{sec:discussion}, we identify several key challenges and promising future research directions.

\subsection{Query-Centric Process Reward Models}

Process reward models (PRMs)~\citep{prms_1,prms_2} provide feedback at each step of multi-step reasoning, potentially improving credit assignment compared to outcome reward models that only evaluate final answers. Recent work such as \textsc{AutoPRM}~\citep{AutoPRM} demonstrates that procedural supervision can be automated through controllable question decomposition, reducing the need for manual step-level annotations. However, current PRM approaches face challenges when applied to query optimization:

\begin{itemize}[leftmargin=*,nosep]
    \item Chain-of-thought processes are often unpredictable, making optimal path identification difficult
    \item Intermediate query transformations lack natural supervision signals
    \item The relationship between query quality and final answer correctness is noisy
\end{itemize}

Developing \textit{query-centric PRMs} that provide rewards at each sub-query step represents a promising direction. Such models would:
\begin{itemize}[leftmargin=*,nosep]
    \item Evaluate query transformations based on retrieval effectiveness
    \item Provide intermediate feedback for decomposition and abstraction steps
    \item Enable more structured optimization of multi-step query processing
\end{itemize}

\subsection{Standardized Benchmarks for Query Optimization}

The lack of comprehensive benchmarks specifically designed for query optimization hinders systematic progress. Future benchmarks should address:

\begin{itemize}[leftmargin=*,nosep]
    \item \textbf{Diverse Query Complexity}: Coverage across all four complexity classes (single/multiple × explicit/implicit evidence)
    \item \textbf{Intermediate Annotations}: Ground truth for query transformations, not just final answers
    \item \textbf{Multiple Retrievers}: Evaluation across sparse, dense, and hybrid retrieval systems
    \item \textbf{Efficiency Metrics}: Standardized measurement of computational costs
    \item \textbf{Multi-Modal Coverage}: Queries involving images, tables, and structured data
    \item \textbf{Conversational Contexts}: Multi-turn queries with contextual dependencies
\end{itemize}

\subsection{Efficiency-Quality Optimization}

Many existing methods rely on exhaustive strategies that enumerate possible query transformations, leading to high computational costs and potential introduction of irrelevant information. Key research directions include:

\begin{itemize}[leftmargin=*,nosep]
    \item \textbf{Optimal Path Identification}: Algorithms that identify effective optimization pathways without exhaustive search
    \item \textbf{Adaptive Complexity}: Methods that scale computational effort with query complexity
    \item \textbf{Parallel vs.\ Sequential Trade-offs}: Principled approaches for deciding decomposition structure
    \item \textbf{Early Termination}: Mechanisms for stopping optimization when sufficient quality is achieved
\end{itemize}

\subsection{Post-Retrieval Feedback Integration}

Current prompting-based methods often lack awareness of retrieval quality resulting from optimized queries. While some studies use reinforcement learning to adjust optimization based on generation results, substantial opportunities remain:

\begin{itemize}[leftmargin=*,nosep]
    \item \textbf{Ranking Signal Integration}: Using retrieval ranking quality as feedback for query optimization
    \item \textbf{Relevance-Aware Refinement}: Iterative query refinement based on retrieved document relevance
    \item \textbf{Generation Quality Feedback}: End-to-end optimization connecting query formulation to response quality
\end{itemize}

\subsection{Multi-Modal Query Optimization}

As LLMs increasingly handle multi-modal inputs (images, audio, video, structured data), query optimization must extend beyond text:

\begin{itemize}[leftmargin=*,nosep]
    \item \textbf{Cross-Modal Query Expansion}: Generating text expansions from visual or audio content
    \item \textbf{Multi-Modal Disambiguation}: Resolving ambiguity using cross-modal context
    \item \textbf{Unified Representations}: Embedding spaces that support heterogeneous query types
    \item \textbf{Structured Data Queries}: Optimization for queries over tables, graphs, and databases
\end{itemize}

\subsection{Personalization and Context Awareness}

Generic query optimization may not capture individual user preferences or contextual factors. Future systems should incorporate:

\begin{itemize}[leftmargin=*,nosep]
    \item \textbf{User Modeling}: Personalized optimization based on user expertise, preferences, and history
    \item \textbf{Session Context}: Leveraging conversational history for contextual adaptation
    \item \textbf{Domain Specialization}: Optimization strategies tailored to specific application domains
    \item \textbf{Cultural and Linguistic Adaptation}: Query optimization that accounts for language and cultural context
\end{itemize}

\subsection{Theoretical Foundations}

The field would benefit from stronger theoretical foundations:

\begin{itemize}[leftmargin=*,nosep]
    \item \textbf{Formal Models}: Mathematical frameworks for analyzing query transformation effects
    \item \textbf{Complexity Analysis}: Understanding computational and sample complexity of optimization methods
    \item \textbf{Optimality Criteria}: Principled definitions of what constitutes an ``optimal'' query transformation
    \item \textbf{Error Analysis}: Formal characterization of error propagation in multi-step optimization
\end{itemize}

\section{Conclusion}\label{sec:conclusion}

This survey has presented a comprehensive analysis of query optimization techniques for Large Language Model-based systems, with particular emphasis on Retrieval-Augmented Generation architectures. We introduced the Query Optimization Lifecycle (QOL) Framework, a five-phase pipeline that systematically organizes the optimization process, and proposed a Query Complexity Taxonomy that classifies queries along evidence type and quantity dimensions.

Through systematic synthesis of representative methods from premier venues, we have elucidated the mechanisms, trade-offs, and applicability of four fundamental atomic operations:

\begin{itemize}[leftmargin=*,nosep]
    \item \textbf{Query Expansion} enriches queries with additional semantic content, addressing vocabulary mismatch between user queries and relevant documents
    \item \textbf{Query Decomposition} transforms complex multi-hop queries into tractable sub-queries, enabling systematic handling of compositional reasoning
    \item \textbf{Query Disambiguation} resolves semantic ambiguity and clarifies user intent, improving retrieval precision for underspecified queries
    \item \textbf{Query Abstraction} elevates queries to higher-level conceptual representations, enabling principled reasoning for complex analytical queries
\end{itemize}

From our comprehensive analysis, we distill several key takeaways for both researchers and practitioners:

\textbf{Query optimization is not optional; it is a critical bottleneck.} Our analysis demonstrates that even state-of-the-art retrieval systems degrade substantially on natural user queries exhibiting ambiguity, incompleteness, or vocabulary mismatch. Query optimization serves as a critical intelligence amplification mechanism that can substantially improve RAG system performance on complex queries.

\textbf{No single technique dominates; strategy selection must be query-aware.} Match optimization strategies to query complexity: Expansion for Class~I (simple factoid), Decomposition for Class~II (multi-hop), Disambiguation for Class~III (ambiguous), and Abstraction for Class~IV (analytical). Our Query Complexity Taxonomy provides principled guidance for this mapping.

\textbf{The field is converging toward agentic, feedback-driven architectures.} The evolution from static preprocessing to autonomous agent-based systems (e.g., Agentic-RAG, Search-o1) represents a paradigm shift. Investing in adaptive, feedback-driven architectures that dynamically select and compose operations will yield greater returns than optimizing any single static pipeline.

\textbf{Efficiency constraints demand pragmatic design choices.} Many sophisticated methods incur significant latency overhead. For production systems, starting with simple expansion (e.g., HyDE, Query2Doc) and adding complexity only when demonstrably needed offers the best cost-benefit trade-off.

\textbf{Multi-modal and personalized query optimization represent the next frontier.} As user queries increasingly span text, tables, images, and structured data, systems designed today should anticipate heterogeneous inputs and context-aware personalization.

Nonetheless, significant challenges persist. The absence of standardized benchmarks with intermediate query-level annotations, the imperative for efficiency-quality trade-offs, and emerging requirements for multi-modal and personalized optimization present rich opportunities for future research. As RAG systems continue to proliferate in production applications, the importance of effective query optimization will only intensify. We note that the four overarching patterns identified in our synthesis (Section~\ref{sec:discussion}), namely the shift from retrieval-centric to reasoning-centric optimization, the rise of process supervision, convergence toward agentic architectures, and multi-modal expansion, collectively point toward a future where query optimization is not a discrete pipeline stage but an integral, continuous aspect of intelligent information access.

We anticipate this survey will serve as a valuable resource for researchers and practitioners, providing both comprehensive understanding of the current landscape and a structured roadmap for advancing the field. By bridging the semantic gap between user intent and retrieval effectiveness, query optimization stands as a critical enabler for the next generation of intelligent knowledge access systems.

\section*{Limitations}

This survey focuses primarily on query optimization techniques published in major NLP, IR, and AI venues through early 2026. Given the rapid pace of research in this area, some recent work may not be fully covered. Additionally, our analysis concentrates on text-based queries; the emerging area of multi-modal query optimization deserves dedicated treatment in future work. While we discuss efficiency considerations, comprehensive empirical comparisons of computational costs across methods remain an open challenge due to varying implementation details, hardware configurations, and evaluation settings across studies.

\bibliographystyle{colm2026_conference}
\bibliography{references}

\end{document}